\documentclass[11pt]{article}

\usepackage{acl}
\usepackage{amssymb}
\usepackage{multirow}
\usepackage{times}
\usepackage{latexsym}
\usepackage{amsmath}

\usepackage[T1]{fontenc}
\usepackage{float}

\usepackage{inconsolata}

\usepackage{graphicx}
\usepackage[capitalise]{cleveref}
\usepackage{url}
\usepackage{booktabs}
\usepackage{bbding}
\usepackage{makecell}
\usepackage{algpseudocode}
\usepackage{amssymb}
\usepackage{pifont} 
\usepackage{amsfonts} 
\usepackage{mathrsfs} 
\usepackage{tabularx}
\usepackage{colortbl}
\usepackage{lscape}
\usepackage{wrapfig}
\usepackage{subfig} 
\usepackage{amsmath}
\usepackage{amssymb}
\usepackage[linesnumbered,ruled]{algorithm2e}

\usepackage{mathptmx} 

\usepackage[english]{babel}
\usepackage{graphicx}
\usepackage[makeroom]{cancel}
\usepackage{hyperref}

\usepackage{microtype}
\usepackage[utf8]{inputenc}
\usepackage[T1]{fontenc}

\usepackage{colortbl}
\usepackage{booktabs}

\usepackage{tcolorbox}
\tcbuselibrary{breakable}

\usepackage{xcolor}

\definecolor{mc-color}{RGB}{245, 250, 255} 
\definecolor{gen-color}{RGB}{247,254,247}
\definecolor{model-color}{RGB}{250, 250, 250}

\definecolor{dark-grey}{RGB}{245,245,245}

\definecolor{white}{RGB}{255,255,255}

\newcommand{\datasetname}{\textsc{BehaviorChain}\xspace}

\title{How Far are LLMs from Being Our Digital Twins? \\A Benchmark for Persona-Based Behavior Chain Simulation}

\author{
    Rui Li$^1$,
    Heming Xia$^2$,
    Xinfeng Yuan$^3$,
    Qingxiu Dong$^1$,
    Lei Sha$^4$,
    Wenjie Li$^2$,
    Zhifang Sui$^1$\thanks{~~Corresponding author} \\
        \normalsize{$^1$State Key Laboratory of Multimedia Information Processing, School of Computer Science, Peking University} \\
        \normalsize{$^2$The Hong Kong Polytechnic University} \quad
       \normalsize{$^3$Fudan University} \quad
        \normalsize{$^4$Beihang University}\\
    \texttt{o\_l1ru1@stu.pku.edu.cn, he-ming.xia@connect.polyu.hk}
}

\begin{document}
\maketitle

\begin{abstract}


Recently, LLMs have garnered increasing attention across academic disciplines for their potential as human \textbf{\textit{digital twins}}, virtual proxies designed to replicate individuals and autonomously perform tasks such as decision-making, problem-solving, and reasoning on their behalf.
However, current evaluations of LLMs primarily emphasize dialogue simulation while overlooking human behavior simulation, which is crucial for digital twins.
To address this gap, we introduce \datasetname, the first benchmark for evaluating LLMs' ability to simulate continuous human behavior.
\datasetname comprises diverse, high-quality, persona-based behavior chains, totaling 15,846 distinct behaviors across 1,001 unique personas, each with detailed history and profile metadata.
For evaluation, we integrate persona metadata into LLMs and employ them to iteratively infer contextually appropriate behaviors within dynamic scenarios provided by \datasetname. Comprehensive evaluation results demonstrated that even state-of-the-art models struggle with accurately simulating continuous human behavior.
Resources are available at \href{https://github.com/O-L1RU1/BehaviorChain}{https://github.com/O-L1RU1/BehaviorChain}.

\end{abstract}

\section{Introduction}
The rapid advancement of large language models (LLMs) has unlocked unprecedented capabilities in artificial intelligence~\cite{ding2023unraveling}, enabling systems to interpret natural language~\cite{lu2024llms}, reason about complex scenarios~\cite{huang2024planning}, and interact with humans through seamless workflow integration~\cite{yang2024llm}.  These advancements have spurred interest across diverse fields in leveraging LLMs to create human \textbf{\textit{digital twins}}—virtual representations of individuals, potentially equipped with personalized data and historical information, allowing them to perform tasks such as decision-making and reasoning on behalf of the individual.

\begin{figure}[t]
    \centering
    \includegraphics[width=1\linewidth]{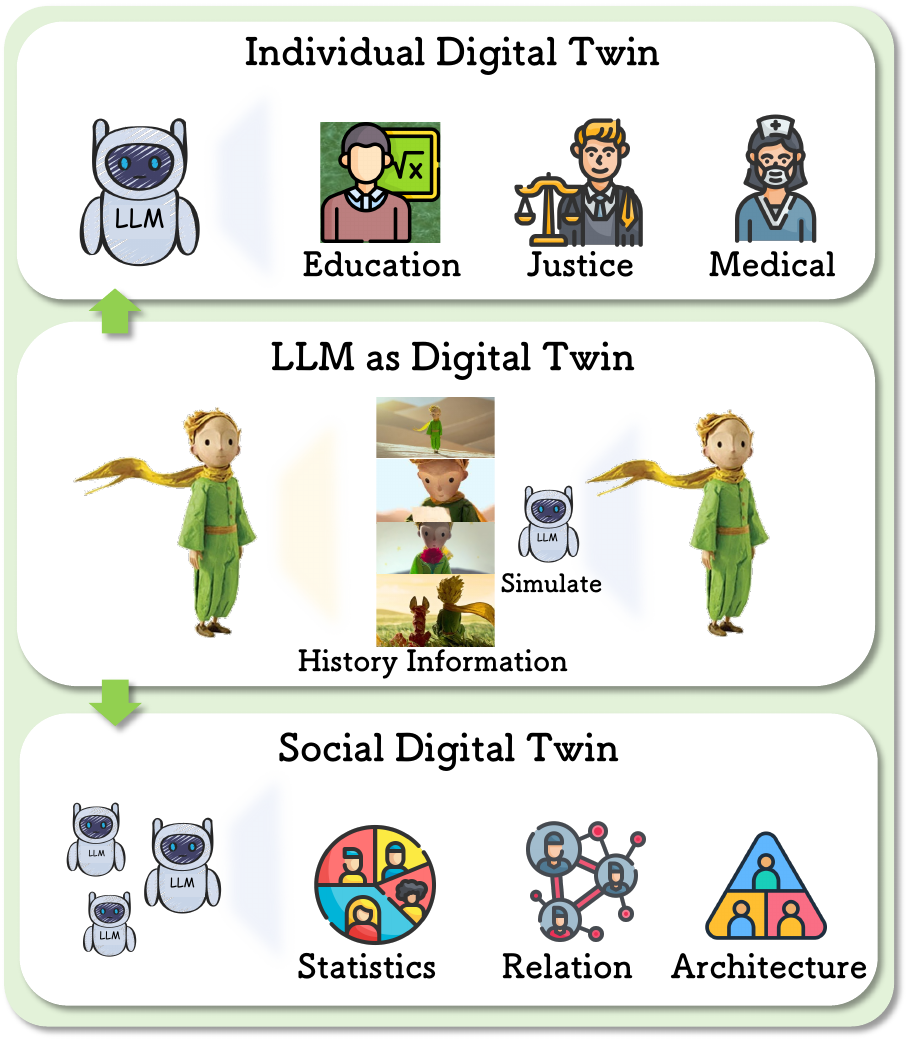}
    \caption{A foundational framework (center) leverages historical data to develop an individual's LLM-powered digital twin.  These individualized digital twins (top) have diverse applications across domains such as healthcare, education, and personalized services.  Furthermore, scaling this approach to collective digital twins (bottom) enables dynamic simulations of societal systems, providing unprecedented insights into emergent behaviors and large-scale social dynamics.}
    \label{fig:application}
\end{figure}

Digital twins, initially conceived for industrial applications such as smart manufacturing~\cite{kreuzer2024artificial}, are increasingly being applied in human-centric domains. 
In fields like personalized healthcare~\cite{wang2024twin} and the metaverse~\cite{lv2022building, aloqaily2022integrating, far2022applying}, these digital replicas of physical entities facilitate predictive scenario testing and real-time monitoring.
The emergence of LLMs with human-like cognition~\cite{gonzalez2024building} and role-playing capabilities~\cite{chen2024from} presents a paradigm shift: these models can synthesize personalized histories to simulate domain-specific expertise~\cite{li2024personal,blasek2023large} (e.g., digital twin teacher), provide emotional companionship~\cite{tu2023characterchat},
and most notably, execute authorized proxy behaviors on behalf of human users.
When scaled to social networks, ensembles of LLM-based twins could further unravel complex human dynamics~\cite{rossetti2024ysocialllmpoweredsocial}, from cultural evolution to crisis response patterns. 
The operational mechanisms and application scenarios of digital twins are illustrated in Figure \ref{fig:application}.

Despite this potential, NLP research aligned with the conceptual framework of LLM-as-digital-twin has predominantly focused on simulating human dialogues~\cite{shen2024roleeval}, knowledge~\cite{shen2024roleeval}, decision-making~\cite{xu2024character} and comprehension ability~\cite{yuan2024evaluating}.
Critically, the simulation of human-like behavior remains underexplored.
Moreover, existing work predominantly relies on static, single-turn simulation tasks or unstable sandbox environments, thus failing to capture LLMs' ability to simulate human continuous behaviors across dynamically evolving contexts—a hallmark of real-world intelligence.
This limitation creates a significant gap between the promise of digital twins and their practical utility in applications demanding longitudinal behavioral fidelity, such as virtual reality avatars or AI-assisted psychotherapy.

To address this gap, we introduce \datasetname, a novel benchmark designed to assess LLMs' ability to simulate continuous human behavior. 
\datasetname comprises 1,001 high-quality, persona-based behavior chains, each containing 10–20 context-behavior nodes, automatically extracted from fiction and biographical literature. Each persona includes detailed profile and history metadata. This literature-based dataset offers a valuable simulation environment for studying continuous human behavior, particularly given the scarcity of real-world behavioral data.

We established an evaluation framework by integrating persona metadata into LLMs and challenging them to progressively recognize or generate persona-based behaviors within dynamically evolving temporal contexts.  
Our evaluation framework comprised two tasks: a multiple-choice behavioral prediction task assessing recognition capabilities, and an open-ended generation task evaluating behavioral synthesis abilities.  

We conducted comprehensive evaluations of ten state-of-the-art LLMs and performed detailed analyses, yielding the following key findings regarding continuous human behavior simulation:
1) Accurately simulating such behavior poses a significant challenge, with even advanced models like GPT-4 achieving sub-60\% accuracy.
2) LLMs are less proficient at simulating non-key behaviors compared to critical ones.
3) A snowball effect occurs during behavior chain completion, where initial errors compound and degrade subsequent predictions. These findings offer valuable insights into the challenges and opportunities in developing LLM-based digital twins.

Our main contributions are as follows:


\begin{itemize}
    \item To bridge the current research gap in digital twin, we propose \datasetname, the first benchmark designed to evaluate LLMs' ability to model behavioral chain dynamics.


    \item 
\datasetname comprises diverse, high-quality behavioral sequences, encompassing 15,846 distinct behaviors across 1,001 unique personas, systematically extracted from literary corpora using an automated, scalable pipeline. 
This dataset offers valuable material for advancing behavior simulation research.



   
    \item
Comprehensive evaluations and analysis of ten state-of-the-art LLMs based on \datasetname revealed that accurately simulating continuous human behavior remains a significant challenge, even for advanced models like GPT-4o.






    \end{itemize}



\section{Related Work}
\label{section_related_work}

\subsection{LLMs as Digital Twins}
Character role-playing serves as a foundation for developing LLMs as digital twins.  Numerous studies have explored adapting LLMs for realistic character simulation through prompt engineering and fine-tuning. For example, Chatharuhi~\cite{li2023chatharuhi} leverages enhanced prompts and character memory banks to mimic the tone, knowledge, and personality of specific anime and film characters within an LLM-based role-playing system. 
Trainable frameworks like Character-LLM~\cite{shao2023characterllm}, CharacterGLM~\cite{zhou2023characterglm}, RoleLLM~\cite{wang2024rolellm} and LD-Agent~\cite{li2024hello} further this approach by fine-tuning LLMs on high-quality character data (e.g., novel texts, screenplay dialogues) to achieve more realistic character simulations.  
Building on these promising role-playing capabilities, evaluations have emerged to assess LLM performance in persona-based dialogue~\cite{shen2024roleeval, wakaki2024comperdial, gao2023livechat, samuel2024personagym}, knowledge recall~\cite{shen2024roleeval}, decision-making~\cite{xu2024character}, and comprehension~\cite{yuan2024evaluating}. 
While not explicitly focused on digital twins, these studies offer valuable insights for the development of digital twins.
Personalized LLM agents~\cite{li2024personal} represent a compelling paradigm of LLMs as digital twins. These agents leverage user historical data to predict needs and provide tailored services~\cite{tan2024personalized,zhuang2406hydra,zhang2024personalized}, assist users with complex professional tasks such as education~\cite{zhang2024simulating}, healthcare~\cite{abbasian2023conversational}, and offer emotional support~\cite{tu2023characterchat}.

\subsection{From Individual to Society}

Individual simulation, focused on replicating individual and group behaviors, forms the cornerstone of LLM-based digital twins. This capability provides the foundation for broader social simulation, where interactions among digital agents create complex networks exhibiting emergent behaviors and societal impacts. Platforms like Y Social \citep{rossetti2024ysocialllmpoweredsocial} exemplify this potential, offering a controlled environment to study online social dynamics, information propagation, and the effects of platform policies through LLM-powered social media simulations.  In economics, LLM-driven agents facilitate both micro-level analysis of pricing strategies and algorithmic collusion risks \citep{fish2024algorithmiccollusionlargelanguage}, and macro-scale simulations of market dynamics \citep{gurcan2024llmaugmentedagentbasedmodellingsocial}.  For population-level modeling, ElectionSim \citep{zhang2024electionsim} pioneers ultra-large-scale simulation through demographically aware voter modeling and customized preference distributions, enabling high-precision election predictions and direct interaction with simulated voter populations.

\begin{figure*}[t]
    \centering
    \includegraphics[width=1\linewidth]{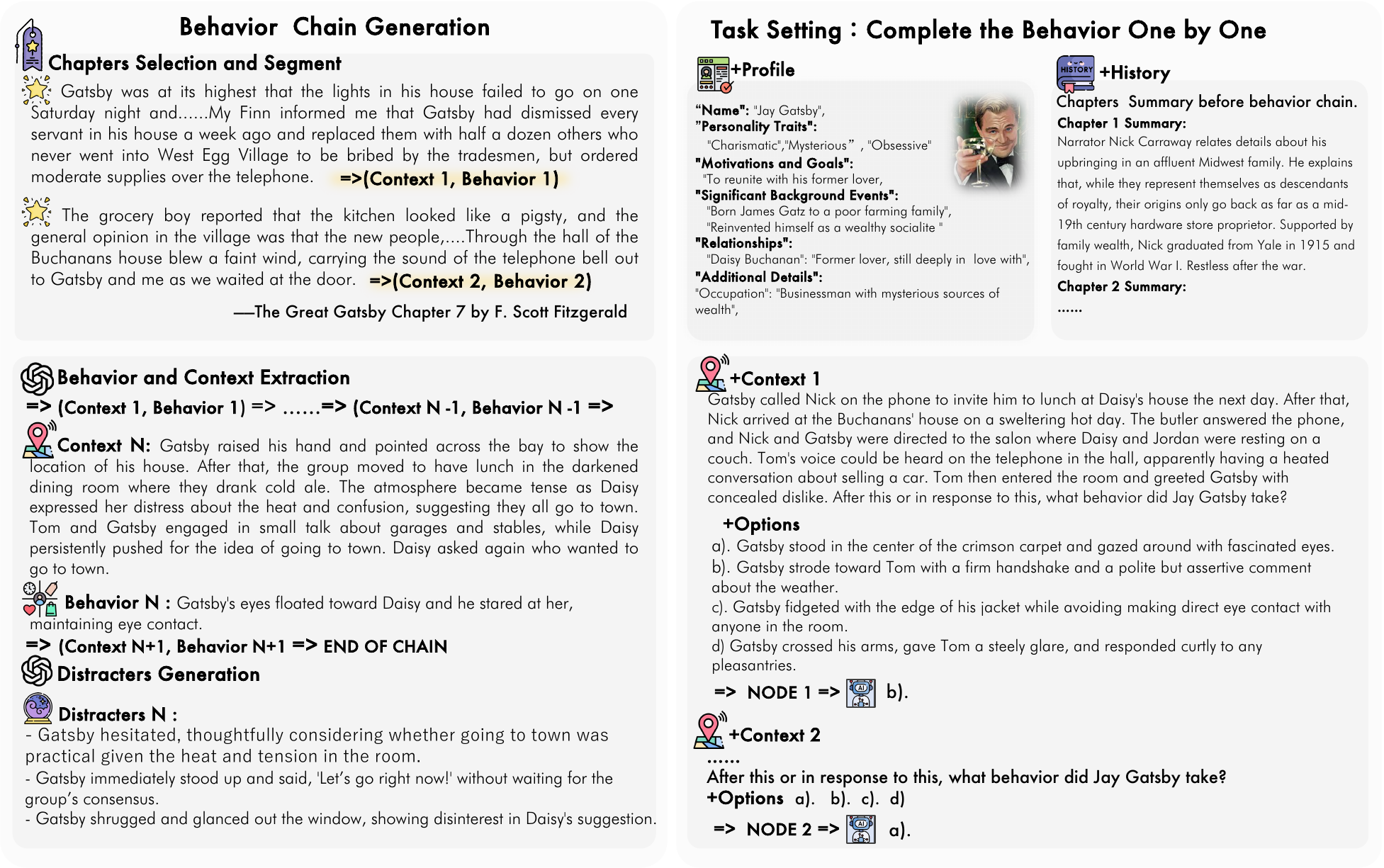}
    \caption{The left image illustrates the process of behavior chain construction. The right image shows our multi-choice task setup, where each node on the behavior chain serves as an individual evaluation. Each input includes persona's profile, history, and all context and ground truth behavior preceding the current node.}
    \label{fig:workflow}
\end{figure*}

\section{Benchmark Crafting}
\label{sec:method}

\label{s:data_construction}
In this section, we introduce the detailed methodology for constructing \datasetname,  focusing on persona-based behavior chain extraction, distractors generation and presenting the multiple-choice and generation task formulations.

\subsection{Overview}

While LLMs offer potential for creating digital twins of individuals, current benchmarks inadequately evaluate their ability to simulate human behavior, especially within dynamic, continuous contexts. To address this, we introduce \datasetname. 
As illustrated in Figure \ref{fig:workflow}, each \datasetname instance is composed of four key components: a persona profile $p$, a historical narrative $h$, a behavior chain $B=\{b_1,b_2,\dots,b_n\}$ of the specific persona, and the contextual setting for each behavior $C=\{c_1,c_2,\dots,c_n\}$.
For the multiple-choice task setup, we 
augment each behavior \( b_i \)
with three distractor behaviors
\( D_i = \{d_1, d_2, d_3\} \).
Example data samples are provided in Appendix \ref{sec:app_instance}.
Historical narratives were constructed by refining chapter summaries from SuperSummary\footnote{\url{https://www.supersummary.com}} using GPT-3.5.
Persona profiles were generated based on character analyses, which were also sourced from SuperSummary. Novels and biographies from Project Gutenberg and purchased e-books served as the source material for extracting high-quality behavior chains.  
In the following sections, we will detail the scalable strategy employed for behavior chain and distractor generation, applicable to any character-centric literature. 
More detailed information and prompts about the dataset construction is provided in Appendix \ref{sec:app_con} and \ref{sec:app_prompt}.

\subsection{Dataset Construction}
The dataset's construction proceeds in three steps: (1) chapter selection and paragraph segmentation based on the frequency of a representative character's presence; (2) paragraph-by-paragraph extraction of behaviors and their corresponding contexts; and (3) distractor generation for each behavior within a given context.


\paragraph{Chapter Selection and Segmentation}

To ensure behavioral continuity and consistency with established character traits, we extracted the main character's behaviors from consecutive chapters in the latter half of each target book.  
Chapters with the highest character name frequencies were selected to maximize the number of behaviors that can be extracted.
These chapters were then evenly divided into paragraphs, from which one behavior per paragraph was extracted.

\paragraph{Behavior and Context Extraction}

This step presents two challenges: the non-uniqueness and meaninglessness of extracted behaviors, and contextual information leakage. 
To address these challenges, we propose a hierarchical context-behavior extraction method based on segmented paragraphs $P = \{p_1, p_2, ..., p_m\}$.
Our framework begins by extracting the first node ${(c_1, b_1)}$ of the chain, where $c_1$ denotes the initial context and  $b_1$ denotes the most key behavior extracted from the first paragraph $p_1$.
For subsequent paragraphs $p_i$ (where $i > 1$), we instruct the model to extract $(c_i, b_i)$ by considering the preceding paragraph $p_{i-1}$, the preceding behavior $b_{i-1}$ , and the current paragraph $p_i$.  
Context-behavior isolation is maintained through post-refinement.
To ensure valid behavior extraction, we use a dual-validation mechanism: similarity and meaningfulness checks.
If either check fails, the current behavior $b_i$ is discarded, and the current paragraph $p_{i}$ is concatenated with the next paragraph $p_{i+1}$ to start a new round of behavior extraction.  
Validated $(c_i, b_i)$ nodes are appended to the chain, and the process continues until all paragraphs are processed or the iteration limit is reached.
Detailed algorithm flow is shown in Algorithm \ref{algo:gen}.


\paragraph{Distractor Generation}
A distractor is a non-optimal option crafted to mislead the target model. For the multiple-choice tasks, we generate three distractors, $D=\{d_1,d_2,d_3\}$, for each extracted context-behavior node $(c_i,b_i)$.
Valid distractors are generated based on two key principles:
\textbf{Context Relevance:} Distractors should be pertinent to the prevailing context. This relevance enhances their deceptiveness, compelling the model to conduct an in-depth analysis of the persona to make the optimal choice.
\textbf{Anchored in Core Character Traits:} Distractors should exhibit substantial deviations from the ground-truth behavior concerning the character's personality traits. This fundamentally guarantees the optimality of the ground truth. 

To adhere to these principles, we instruct the model to first identify the core personality traits reflected in the original behavior $b_i$, then generate three context-constrained adversarial distractors, each exhibiting a distinct personality trait. 



\paragraph{Manual Examination}

To ensure data quality, we recruited ten native English-speaking undergraduate annotators, compensating them at the regional minimum wage. These annotators assessed the meaningfulness of each behavior and the logical coherence between the context and the behavior, flagging any low-quality entries.
After iterative refinement, we curated 1,001 behavior chains of fictional and non-fictional characters, each uniquely mapped to its source monograph (with no cross-book duplicates).
More details regarding the human annotation process can be found in Appendix \ref{app:annotation}. The dataset statistics can be found in \ref{app:data_analysis}.



\begin{algorithm}[t]
\caption{Chain Generation}
\label{algo:gen}
\SetKwData{Initialization}{\textbf{Initialization}}
\DontPrintSemicolon
\SetAlgoLined
\KwIn {Paragraph Set $P=\{p_1,p_2,\dots,p_m\}$, Iteration Count $I$}
\KwOut {chain $\{(c_1,b_1),(c_2,b_2),\dots,(c_n,b_n)\}$}
\Initialization \\
$c_1, b_1 \gets \text{Generate\_First\_Chain\_Node}(p_1)$ \\
$chain \gets \{(c_1, b_1)\}$ \\
$i \gets 2$ \\
\While{$i \leq m$}{
    $c_i, b_i \gets \text{Generate\_Next\_Chain\_Node}(p_{i-1}, p_i, b_{i-1})$ \\
    $s_i \gets \text{is\_Similar}(b_i, b_{i-1})$ \\
    $m_i \gets \text{is\_Meaningful}(b_i)$ \\
    \If{$s_i = 1 \textbf{ or } m_i = 1$}{
        $p_i \gets \text{Concatenate}(p_{i-1}, p_i)$ \\
        \If{$i + 1 \leq m$}{ 
            $P \gets P \setminus p_i$ \\
            $m \gets m - 1$ \\
        }
        \textbf{Continue} \\
    }
    $c_i \gets \text{Refine\_for\_Isolation}(c_i)$ \\
    $chain \gets chain \cup \{(c_i, b_i)\}$ \\
    $i \gets i + 1$ \\
    \If{$i \geq I$}{
        \textbf{Break}
    }
}
\Return {chain}
\end{algorithm}

\subsection{Task Formulation}
\label{task_formulation}


We designed multiple-choice and generation tasks to evaluate LLMs' ability to simulate persona-based continuous behavior in closed- and open-domain settings.

\paragraph{Multiple-choice Task}
The input to the LLM is defined as \( x_i = (p, h, \text{chain}_{<i}, c_i, O) \), where \( \text{chain}_{<i} \) denotes the historical sequence of context-behavior pairs \(\{(c_1, b_1), (c_2, b_2), \dots, (c_{i-1}, b_{i-1})\}\), and \( O \) comprises the correct answer \( b_i \) alongside a set of highly plausible distractors \( D = \{d_1, d_2, d_3\} \). The model is tasked with identifying \( b_i \) from \( O \). This multiple-choice framework provides a structured mechanism to evaluate the LLM’s capacity for fine-grained, persona-based behavioral reasoning and its ability to synthesize information across spatiotemporal-evolving contexts.

\paragraph{Generation task} 
In contrast to the multiple-choice format, the generation task eliminates predefined options, affording the model greater flexibility to produce contextually relevant behaviors. For a given input \( x_i = (p, h, \text{chain}_{<i}, c_i) \), the model is tasked with generating a behavior \( y \) conditioned on the current context. Rather than relying on strict matching between generated outputs \( y \) and ground-truth behaviors \( b_i \), we employ ChatGPT-4o-latest to assess whether \( y \) aligns with the contextual constraints and the persona's predefined personality traits. This task evaluates the model’s capacity to synthesize open-domain behaviors, thereby more closely mirroring real-world applications where future digital twins are anticipated to operate autonomously, generating contextually appropriate behaviors rather than selecting from a finite set of predefined choices.

\section{Experiments}
\label{sec:exp}

\begin{table*}[t]
\centering
\newcolumntype{C}{>{\columncolor{mc-color}}c}
\newcolumntype{G}{>{\columncolor{gen-color}}c}
\resizebox{1\textwidth}{!}{
\begin{tabular}{lCGCGCGCGCGCGCG}

\toprule
& \multicolumn{4}{c}{\textbf{Fictional Persona}} & \multicolumn{4}{c}{\textbf{Nonfiction Persona}} & \multicolumn{4}{c}{\textbf{All}} \\ \cmidrule(lr){2-5} \cmidrule(lr){6-9} \cmidrule(lr){10-13}
& \multicolumn{2}{c}{\textbf{Multiple-choice}} & \multicolumn{2}{c}{\textbf{Generation}} & \multicolumn{2}{c}{\textbf{Multiple-choice}} & \multicolumn{2}{c}{\textbf{Generation}}& \multicolumn{2}{c}{\textbf{Multiple-choice}} & \multicolumn{2}{c}{\textbf{Generation}} \\ \cmidrule(lr){2-3} \cmidrule(lr){4-5} \cmidrule(lr){6-7} \cmidrule(lr){8-9} \cmidrule(lr){10-11} \cmidrule(lr){12-13}
\cellcolor{model-color}\textbf{Model} & \cellcolor{white}\textbf{AvgScore} & \cellcolor{white}\textbf{CumScore} & \cellcolor{white}\textbf{AvgScore} & \cellcolor{white}\textbf{CumScore} & \cellcolor{white}\textbf{AvgScore} & \cellcolor{white}\textbf{CumScore} & \cellcolor{white}\textbf{AvgScore} &
\cellcolor{white}\textbf{CumScore}&
\cellcolor{white}\textbf{AvgScore} & \cellcolor{white}\textbf{CumScore} & \cellcolor{white}\textbf{AvgScore} & \cellcolor{white}\textbf{CumScore} \\ \midrule
\multicolumn{13}{c}{\cellcolor{dark-grey}\textbf{Closed-Source Models}} \\ \midrule 
\cellcolor{model-color}\textbf{gpt-3.5} & 0.276 & 0.047 & 0.273 & \textbf{0.056} & 0.328 & 0.066 & 0.366 & 0.120 & 0.302 & 0.057 & 0.320 & 0.088 \\
\cellcolor{model-color}\textbf{gpt-4o} & \textbf{0.515} & \textbf{0.128} & \textbf{0.366} & \textbf{0.082} & \textbf{0.602} & \textbf{0.188} & \textbf{0.575} & \textbf{0.297} & \textbf{0.559} & \textbf{0.158} & \textbf{0.471} & \textbf{0.189} \\
\cellcolor{model-color}\textbf{claude-3.5-haiku} & 0.433 & 0.094 & 0.133 & 0.025 & 0.497 & 0.116 & 0.323 & 0.117 & 0.465 & 0.105 & 0.228 & 0.071 \\
\cellcolor{model-color}\textbf{qwen-plus} & 0.435 & 0.094 & 0.093 & 0.015 & 0.520 & 0.135 & 0.206 & 0.048 & 0.478 & 0.115 & 0.149 & 0.032 \\

\multicolumn{13}{c}{\cellcolor{dark-grey}\textbf{Open-Source Models}} \\ \midrule 
\cellcolor{model-color}\textbf{deepseek-chat} & 0.474 & 0.109 & 0.094 & 0.014 & 0.560 & 0.154 & 0.220 & 0.062 & 0.517 & 0.132 & 0.157 & 0.038 \\
\cellcolor{model-color}\textbf{llama-3.1-8b} & 0.331 & 0.059 & 0.132 & 0.028 & 0.350 & 0.068 & 0.129 & 0.022 & 0.341 & 0.064 & 0.131 & 0.025 \\
\cellcolor{model-color}\textbf{llama-3.1-70b} & \textbf{0.531} & \textbf{0.137} & 0.221 & 0.038 & \textbf{0.617} & \textbf{0.206} & 0.326 & 0.068 & \textbf{0.574} & \textbf{0.172} & 0.274 & 0.053 \\
\cellcolor{model-color}\textbf{mixtral-8x7b} & 0.159 & 0.025 & 0.043 & 0.006 & 0.200 & 0.035 & 0.252 & 0.070 & 0.179 & 0.030 & 0.148 & 0.038 \\
\cellcolor{model-color}\textbf{qwen2.5-72b} & 0.403 & 0.082 & \textbf{0.230} & 0.044 & 0.479 & 0.114 & \textbf{0.420} & \textbf{0.151} & 0.441 & 0.098 & \textbf{0.325} & \textbf{0.098} \\
\cellcolor{model-color}\textbf{yi-1.5-34b} & 0.274 & 0.049 & 0.031 & 0.006 & 0.292 & 0.062 & 0.293 & 0.089 & 0.283 & 0.056 & 0.162 & 0.048 \\
\bottomrule
\end{tabular}
}
\label{tab:main}
\caption{Models performance on \datasetname, showing average AvgScore and CumScore results for simulating fictional and nonfictional characters' behaviors in multiple-choice and generation settings, along with overall average AvgScore and CumScore.}
\end{table*}

\subsection{Experiments Setup}

\subsubsection{Models}

We selected ten LLMs for our experiments, encompassing closed-source models Claude-3.5-Haiku-20241022, DeepSeek-Chat~\cite{deepseekai2025deepseekr1incentivizingreasoningcapability}, GPT-3.5-0613, GPT-4o-2024-11-20~\citep{openai2024gpt4o} and Qwen-Plus-Latest, and open-source models consisting of Meta-Llama-3.1-8B-Instruct~\citep{dubey2024llama}, Meta-Llama-3.1-70B-Instruct, Mixtral-8x7B-Instruct-v0.1~\citep{jiang2023mixtral}, Qwen2.5-72B-Instruct~\citep{qwen2025qwen25technicalreport}, and Yi-1.5-34B-Instruct~\citep{yi2023yi}.

\subsubsection{Evaluation Metrics}
\label{metric}


To quantify LLM performance in persona-based behavior chain simulation, we introduce two persona-level metrics: Average Chain Score (AvgScore) and Normalized Cumulative Chain Score (CumScore). Overall benchmark performance is then computed by aggregating persona-level scores using expectation.


\paragraph{Average Chain Score (AvgScore)} AvgScore measures a model's ability to capture discrete behaviors by calculating the node-wise accuracy of behavior identification or generation within a chain of length $n$. 
$A_i$ denotes the correctness of the model's response at node $i$. 
The criteria for correctness vary by task (see Section \ref{task_formulation}).
\begin{equation}
\text{AvgScore} = \frac{1}{n}\sum_{i=1}^n A_i
\end{equation}

\paragraph{Normalized Cumulative Chain Score (CumScore)} 
CumScore measures models' higher-order ability to continuously capture behaviors consistent with both persona and context throughout a behavior chain.
CumScore is calculated by summing the lengths of all m consecutively correct behavioral sequences within the chain and then normalizing the result.
\( L_k \) represents the length of the \( k \)-th consecutively correct sequence, and \( n \) represents the total number of nodes.
\begin{equation}
\text{CumScore}_{\text{norm}} = \frac{\sum_{k=1}^m L_k}{\sum_{i=1}^n i}
\end{equation}

\begin{table}[h]
    \centering
    \resizebox{0.4\textwidth}{!}{%
        \begin{tabular}{lcc}
            \toprule
            & \textbf{AvgScore} & \textbf{CumScore} \\
            \midrule
            \textbf{Random Sample} & 0.2471 & 0.0400 \\
            \textbf{Half Sample}  & 0.4956 & 0.1138 \\
            \bottomrule
        \end{tabular}%
    }
    \caption{Reasonable Baseline AvgScore and CumScore obtained through random and half selection in the multiple-choice task.}
    \label{tab:baselines}
\end{table}

\paragraph{Reasonable Baseline Metric Scores}
Both AvgScore and CumScore range from 0 to 1, where higher values indicate better model performance. To establish meaningful baselines for our multiple-choice task, we present the average AvgScore and CumScore obtained through random selection and half selection (representing a scenario where there is a 50\% probability of making the correct choice at each node) within the benchmark, as shown in Table \ref{tab:baselines}. 
These baselines can serve as references: metric scores falling below the random selection baseline signify exceptionally poor performance, while scores exceeding the half-selection baseline can be considered a passing grade. 
Furthermore, we can draw parallels to our generation tasks by referencing baseline statistics derived from these multiple-choice experiments. 
It is important to note that the inherently lower CumScore arises from its stringent requirement for consecutive correct predictions.



\subsection{Main Result}

Based on \datasetname, we systematically evaluated the behavior chain simulation capabilities of current leading LLMs across \textit{multiple-choice} and \textit{generation} tasks. 
More experimental details are provided in Appendix \ref{sec:app_exp}.

\paragraph{Multiple-choice vs Generation}
Models generally perform better in the multiple-choice task compared to the generation task, with AvgScore ranging from 0.159 to 0.474. 
For example, GPT-4o achieves AvgScore of 0.515 in the multiple-choice task, but its score drops to 0.366 in generation for fictional settings.
These results highlight that models struggle more with generating consistent, contextually accurate behaviors across dynamic contexts in open domains.

\begin{figure}[htbp]
    \centering
    \includegraphics[width=1\linewidth]{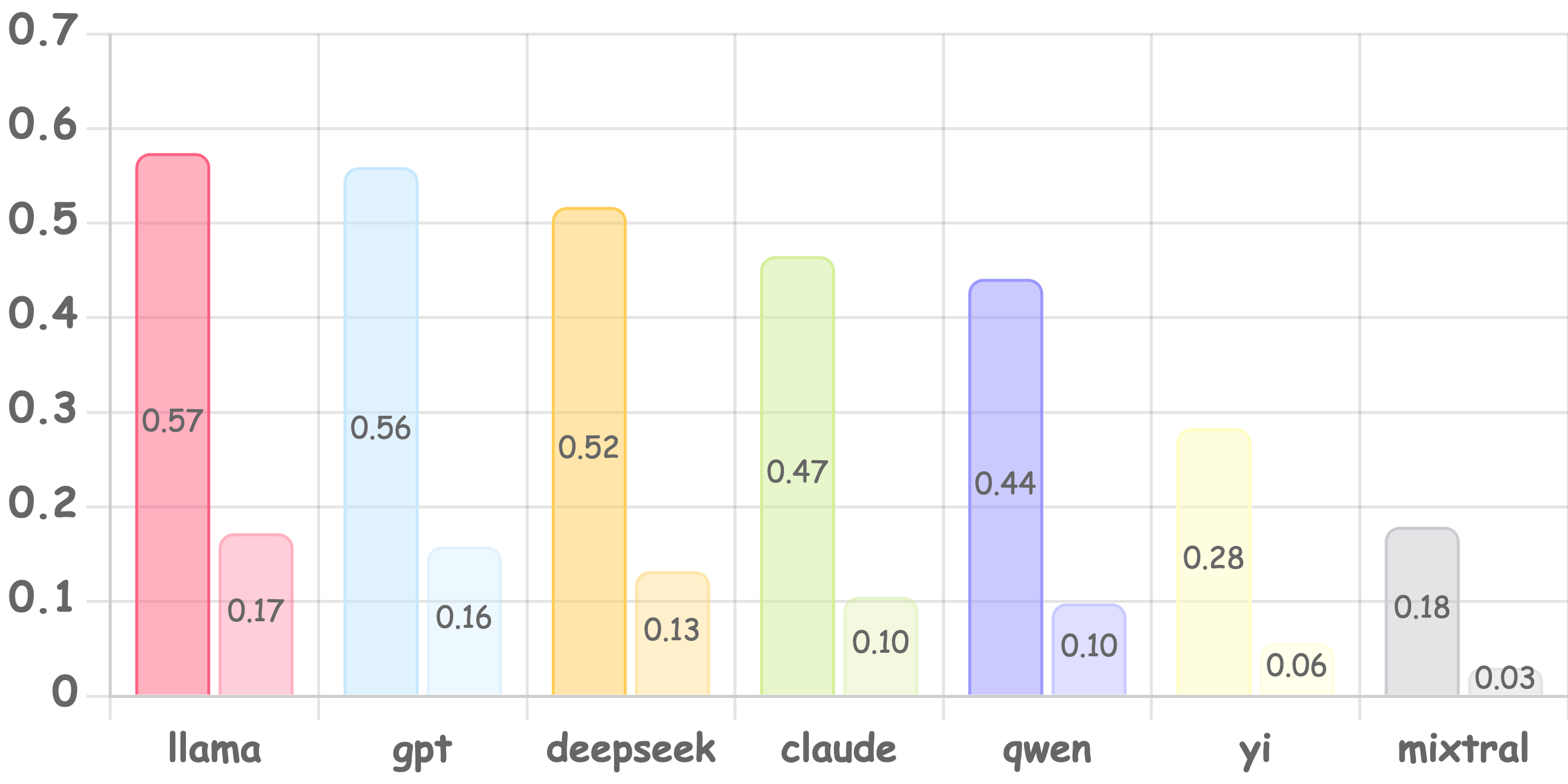}
    \caption{Different model family's average AvgScore (dark bars) and CumScore (light bars) on \datasetname.}
    \label{fig:models}
\end{figure}

\paragraph{Model Family Comparison} 
As shown in Figure \ref{fig:models}, a cross-model comparison reveals that leading models like GPT-4o and Llama-3.1-70B achieve AvgScore of 0.515 and 0.531, respectively, in the multiple-choice task, significantly outperforming other models (ranging from 0.159 to 0.474). This trend aligns with the general capabilities of existing models. In the generation task, GPT-4o also maintains its advantage, outperforming other models by at least 5\%–30\% in simulating fictional personas and by 10\%–40\% in simulating nonfiction personas.

\begin{table}[htbp]
    \centering
    \resizebox{0.5\textwidth}{!}{%
        \begin{tabular}{lcccccc}
            \toprule
            \multirow{2}{*}{Model} & \multicolumn{2}{c}{\textbf{Fictional}} & \multicolumn{2}{c}{\textbf{Nonfictional}} & \multicolumn{2}{c}{\textbf{All}} \\
            \cmidrule(lr){2-3} \cmidrule(lr){4-5} \cmidrule(lr){6-7}
            & \textbf{AvgScore} & \textbf{CumScore} & \textbf{AvgScore} & \textbf{CumScore} & \textbf{AvgScore} & \textbf{CumScore} \\
            \midrule
            \rowcolor[gray]{0.99} \textbf{llama-3.2-1b} & 0.210 & 0.032 & 0.231 & 0.037 & 0.214 & 0.033 \\
            \rowcolor[gray]{0.95} \textbf{llama-3.1-8b} & 0.331 & 0.059 & 0.350 & 0.068 & 0.341 & 0.064 \\
            \rowcolor[gray]{0.90} \textbf{llama-3.1-70b} & 0.531 & 0.137 & 0.617 & 0.206 & 0.574 & 0.172 \\
            \rowcolor[HTML]{fffbf5} \textbf{qwen-2.5-14b} & 0.359 & 0.067 & 0.428 & 0.098 & 0.374 & 0.074 \\
            \rowcolor[HTML]{fff2eb} \textbf{qwen-2.5-72b} & 0.403 & 0.082 & 0.479 & 0.114 & 0.441 & 0.098 \\
            \rowcolor[HTML]{f9ffff} \textbf{yi-1.5-9b} & 0.213 & 0.034 & 0.242 & 0.046 & 0.219 & 0.037 \\
            \rowcolor[HTML]{ebf8ff} \textbf{yi-1.5-34b} & 0.274 & 0.049 & 0.292 & 0.062 & 0.283 & 0.056 \\
            \bottomrule
        \end{tabular}%
    }
    \caption{Scaling performance on BehaviorChain across model families and sizes (llama-3: 1b, 8b, 70b; qwen-2.5: 14b, 72b; yi-1.5: 9b, 34b), with darker shades indicating stronger performance within each family.}
    \label{tab:scale}
\end{table}

\paragraph{Scaling Laws in Behavior Simulation}
The positive correlation between model scale and behavior chain simulation performance is clear, particularly within the Llama-3 family.
For instance, the Llama-3.1-70B outperforms its 8B parameter counterpart in nonfiction setting by a substantial 76.5\% in AvgScore (0.617 vs. 0.350) and an impressive 205\% in CumScore (0.206 vs. 0.068).
To further illustrate the scaling law, we have expanded our evaluation on the BehaviorChain benchmark to include several model families and models of varying sizes. Specifically, we present results for the following groups: Llama-3.2-1B-instruct, Llama-3.1-8B-instruct, and Llama-3.1-70B-instruct; Qwen-2.5-14B-instruct and Qwen-2.5-72B-instruct; as well as Yi-1.5-9B-instruct and Yi-1.5-34B-instruct. As demonstrated in Table \ref{tab:scale}, the performance of all these models aligns with the expected scaling laws.





\paragraph{Nonfiction vs Fiction}
Performance varies notably between nonfiction and fiction settings. 
For example, Qwen-plus achieves a 0.194 higher average accuracy simulating nonfiction personas (0.520) compared to fiction personas (0.435). 
This performance difference is generally observed across models. 
This disparity likely stems from two key factors: (i) the stronger causal links in biographical narratives, facilitating deterministic reasoning in the behavior chain, and (ii) the unconventional psychological motivations of fictional personas, requiring more creative inference.

\paragraph{Models Struggle with Simulating Behavior Chains}
Even the top-performing models, the closed-source GPT-4o and the open-source Llama-3.1-70b-instruct, achieve average AvgScore below 0.62 and average CumScore below 0.21 for the multiple-choice task, with even lower performance in the generation task. 
The low scores demonstrate that existing models still struggle with accurately simulating person-based behavior chains, particularly in maintaining long-range consistency across these chains.



\subsection{Analysis}
\label{sec:analysis}

\subsubsection{Key Behavior vs Sub-Key Behavior}

\begin{figure}
    \centering
    \includegraphics[width=1\linewidth]{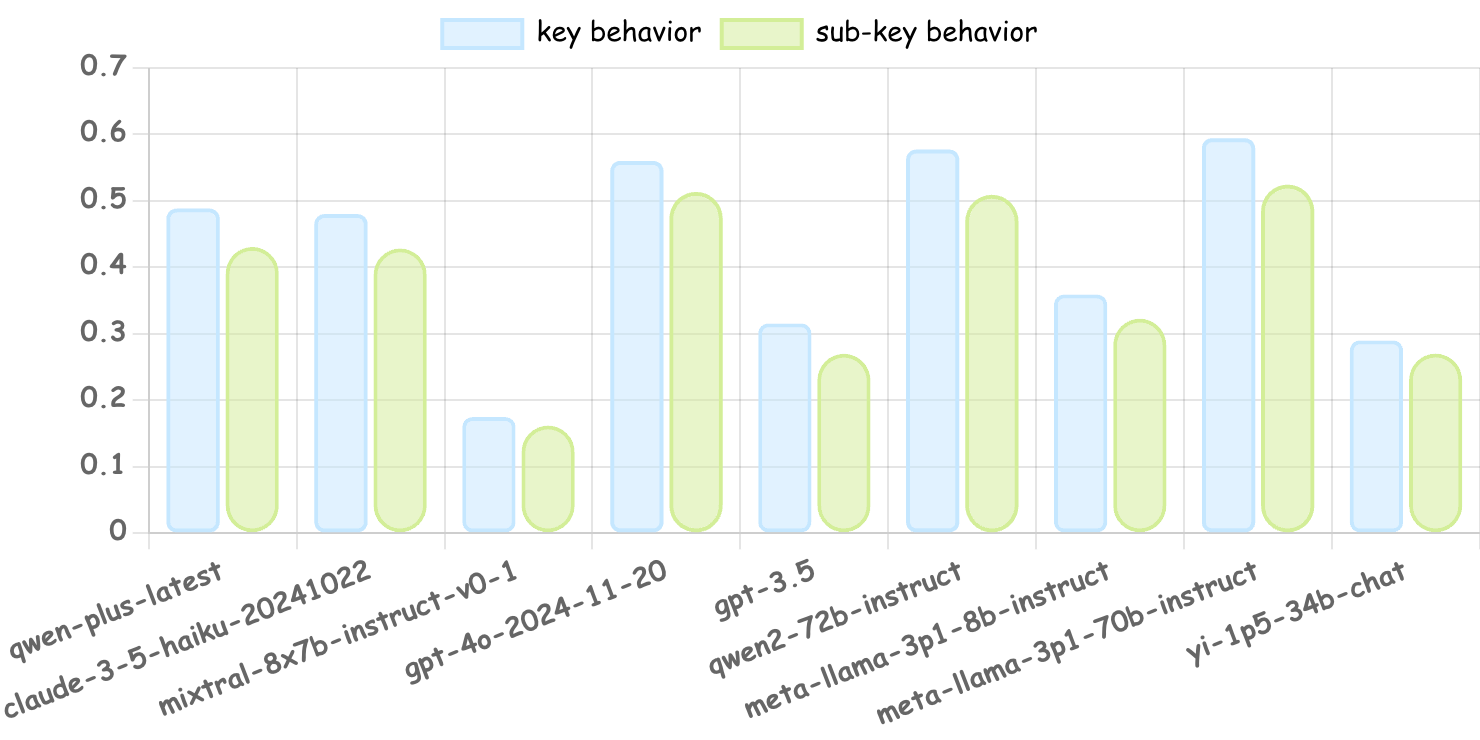}
    \caption{AvgScore performance across the entire dataset for multiple models at key behavior and sub-key behavior.}
    \label{fig:level}
\end{figure}

Not all behaviors are equally important.
To investigate the models' ability to capture key behaviors, we categorized behaviors based on their presence in the chapter summaries collected from SuperSummary.
Each behavior chain was thus divided into sets of key nodes $B_k$ and sub-key nodes $B_{sub}$, yielding a 3:7 ratio.
We then examined the models' AvgScore on these two sets. 
Figure \ref{fig:level} shows that models prioritize capturing important behaviors, as all models achieved higher AvgScore scores on key behaviors than on sub-key behaviors in both fiction and nonfiction settings.
This also reveals models' shortcomings in simulating sub-key behaviors.


\subsubsection{History Length \& Context Length}

\paragraph{History Length}

\begin{figure}
    \centering
    \includegraphics[width=1\linewidth]{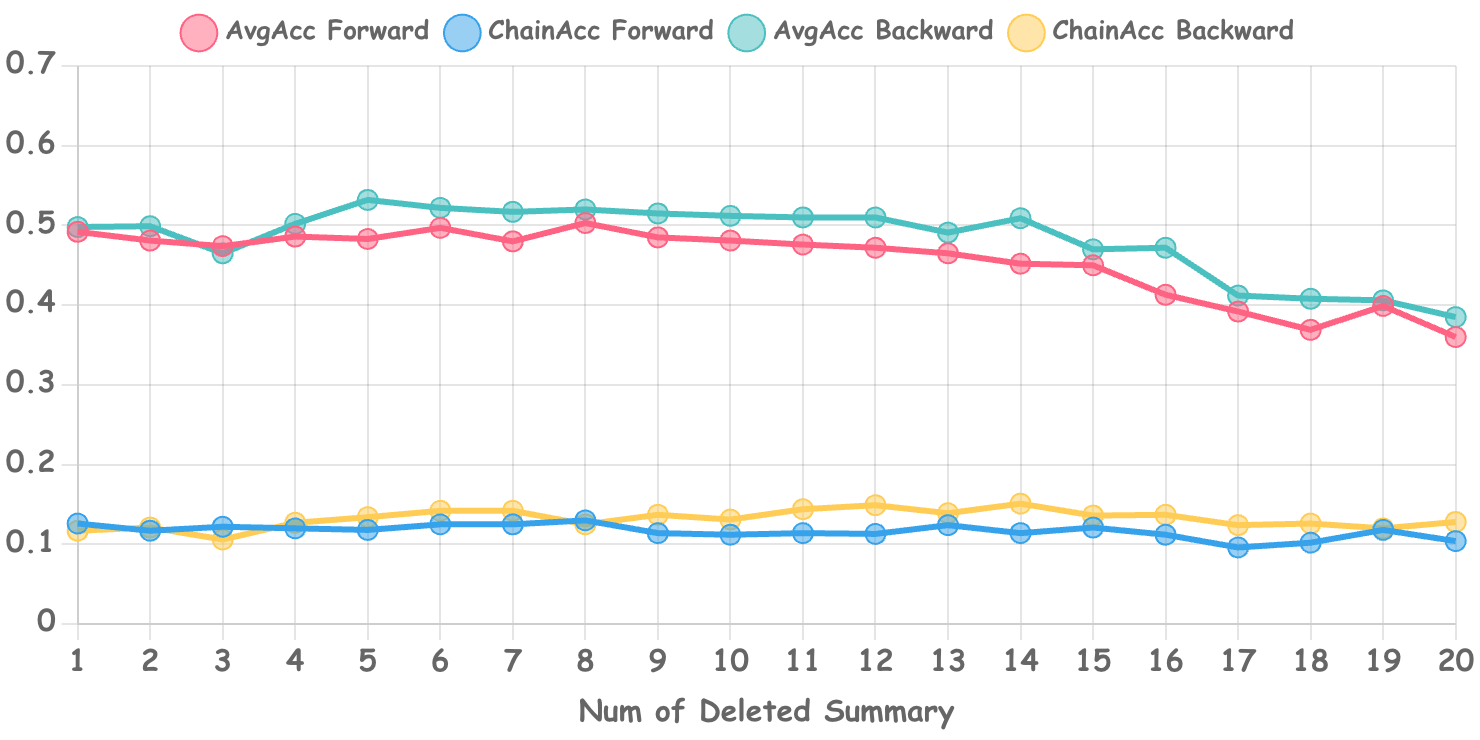}
    \caption{Impact of number of deleted summaries on AvgScore and CumScore (forward and backward directions).}
    \label{fig:length}
\end{figure}

Using chapter summaries as the unit of history, we investigate the impact of history completeness by truncating history summaries both chronologically (forward) and in reverse (backward).
\textit{Forward truncation} progressively removes the earliest history, while \textit{backward truncation} removes the most recent.
We calculated AvgScore and CumScore for each truncation strategy.

As shown in Figure \ref{fig:length}, backward truncation leads to a gradual decline in AvgScore, suggesting that recent historical information contributes crucial contextual information for predicting subsequent behaviors. Forward truncation exhibits a more resilient performance, with AvgScore remaining relatively stable with the removal of early history. 
This implies that the core narrative thread and contextual cues are often established lately.
However, beyond a certain threshold of removed recent history (around 10 chapters' summaries in this average case), we observe a more pronounced drop in AvgScore performance. This indicates that a significant absence of character history drastically impairs LLMs' ability to accurately simulate character behaviors.

\paragraph{Context Length}

\begin{table}
\centering
\begin{tabular}{lcc}
\toprule
\textbf{NODE ID} & \textbf{AvgScore} & \textbf{CumScore} \\
\midrule
\textbf{$\#$1 - $\#$5} & 0.402 & 0.210 \\
\textbf{$\#$6 - $\#$10} & 0.440 & 0.243 \\
\textbf{$\#$11 - $\#$15} & 0.416 & 0.229 \\
\textbf{$\#$16 - $\#$20} & 0.405 & 0.213 \\
\bottomrule
\end{tabular}
\caption{AvgScore and CumScore at different stages of behavior chains.}
\label{tab:node}
\end{table}

We investigated the impact of context length on model performance to explore whether LLMs exhibit a demonstration enhancement akin to in-context learning (ICL). To this end, we selected 20-step behavior chains and temporally segmented them into four phases. We then examined the distribution of correct behavior nodes. As shown in Table \ref{tab:node}, the distribution of correct behavior nodes exhibited a peak between nodes 6 and 10. This localized performance peak may indicate a form of short-term in-context learning or a priming effect, suggesting the model is better equipped to handle behavior prediction within this middle window.  However, this peak is not sustained throughout the entire chain, indicating that accurately simulating character behaviors becomes increasingly difficult as the chain unfolds.

\subsubsection{Snowball Effect in the Behavior Chain}


\begin{table}
    \centering
    \resizebox{0.45\textwidth}{!}{
    \begin{tabular}{lcc}
        \toprule
        \textbf{History Behavior Type} & \textbf{AvgScore} & \textbf{CumScore} \\
        \midrule
        \textbf{Ground Truth} & 0.528 & 0.143 \\
        \textbf{Chosen Behavior} & 0.420 & 0.100 \\
        \bottomrule
    \end{tabular}%
    }
    \caption{Comparison of model performance on AvgScore and CumScore with ground truth behavior vs. chosen behavior history}
    \label{tab:my_label}
\end{table}

We examined the snowball effect of incorrect behavior correction on subsequent simulation in behavior chains using a multiple-choice task setup.
At $i$-th node, the model received input $x_i = (p, h, chain_{<i}, c_i, O)$, where $chain_{<i} = \{(c_1, b_1), (c_2, b_2), \dots, (c_{i-1}, b_{i-1})\}$ represents the preceding context and behavior, and $O$ contains the correct behavior $b_i$ and distractors ${d_1, d_2, d_3}$. 
The model is tasked with selecting the next behavior, denoted as $\hat{b}_i$.
To simulate the snowball effect of incorrect choices, we iteratively replaced the ground truth behavior $b_i$ with the chosen behavior $\hat{b}_i$ at each node $i$ throughout the chain.
As a result, for subsequent nodes, the $chain_{<i}$ in input $x_i$ becomes $\{(c_1, \hat{b}_{1}), (c_2, \hat{b}_{2}), \dots, (c_{i-1}, \hat{b}_{{i-1}})\}$.  
As shown in Table \ref{tab:my_label}, by comparing the model's AvgScore and CumScore under both the normal (ground truth) and replacement settings, we observed a substantial performance decline in the latter.
This confirms the accumulation of errors, where incorrect behavior choices at earlier nodes exacerbate the likelihood of errors at later nodes.
This snowball effect suggests that in more realistic scenarios, where ground-truth is unavailable, simulating continuous behavior becomes more challenging. Using the model's previous behavior records as history significantly impacts subsequent behavior simulation, causing the model to deviate further from the character.

\subsubsection{Temporal Bias}



\begin{figure}[t]
    \centering
    \includegraphics[width=1\linewidth]{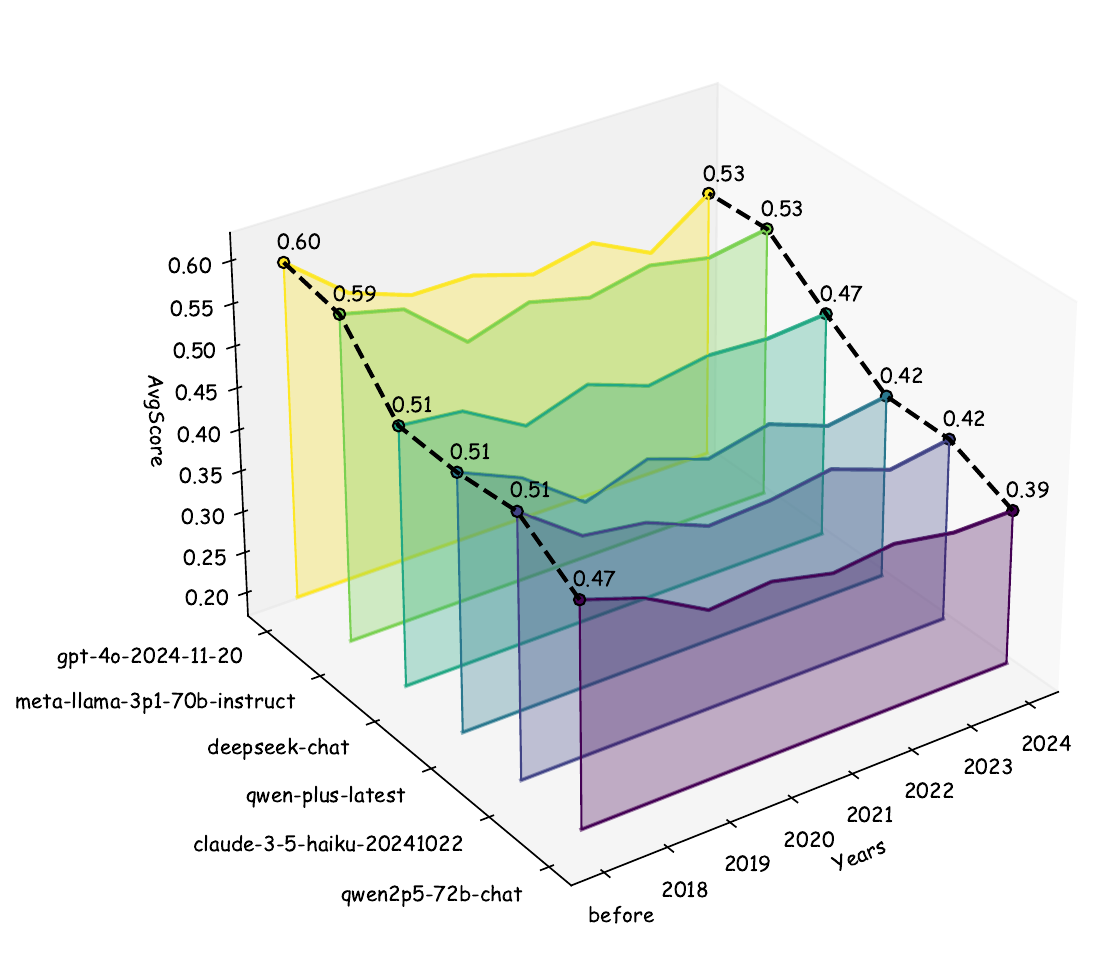}
    \caption{AvgScore (z-axis) for various models (x-axis) plotted against the publication year of the works used for behavior extraction (y-axis).}
    \label{fig:year}
\end{figure}



We categorized \datasetname by the publication year of its source literature (Figure \ref{fig:bing}). We observed that model performance decreases as publication year increases.
As shown in Figure \ref{fig:year}, all models performed better on characters from older publications than on those from more recent ones. 
This is likely due to pre-training data bias, as LLMs are typically trained on a large corpus of classic texts, with relatively fewer contemporary works, contributing to the models' stronger grasp of earlier character behaviors. Therefore, we propose the use of recent literature, specifically works published in 2024, as a hard set for evaluation.

Further results analysis is provided in Appendix \ref{sec:app_result}.


\section{Conclusion}

We introduced \datasetname, a novel benchmark for evaluating LLMs' ability to simulate continuous human behavior.  Comprising 1,001 persona-based behavior chains extracted from literature, it addresses a critical gap in LLM assessment, particularly given the scarcity of real-world behavioral data.  Our evaluation framework, encompassing behavioral recognition and generation tasks, revealed significant challenges for ten state-of-the-art LLMs.
Further experiments analyzing the influence of key behaviors, history, and context characteristics provided valuable insights into the factors that contribute to or hinder successful persona-based behavior simulation.
\datasetname provides a valuable resource for developing robust digital twins.

\section*{Limitations}


\datasetname focuses on English-language Western literature, lacking representation of non-Western cultural norms and behavioral expressions. This cultural narrowness limits applicability to global digital twin deployments.
Generation task evaluation via GPT-4o introduces model bias, as the assessor’s cultural/personal biases may influence consistency judgments. Human evaluation, though more reliable, was limited to data validation rather than comprehensive scoring.

Our current dataset, while encompassing literary works from diverse cultural and national backgrounds, is presently limited to English-language materials. We recognize the critical importance of investigating digital twin applications across multilingual contexts. In subsequent phases of this research, we plan to expand the dataset to incorporate multiple languages, thereby enabling comprehensive evaluation of models' capabilities in simulating human behavior through multilingual interactions.
In the current study, we have focused exclusively on data construction and model evaluation while thoroughly examining the model's limitations across multiple dimensions. The aspect of enhancing the model's capability for continuous behavioral simulation remains underexplored, and we plan to address in subsequent research. 


\section*{Ethics Statement}
This paper introduces \datasetname, a benchmark for evaluating LLMs' ability to simulate human behavior, raising several ethical considerations.  While the benchmark itself doesn't directly generate content that could be harmful, the use of LLMs for behavior simulation has the potential for misuse, such as creating deceptive or manipulative content. The dataset construction process, which involves extracting behaviors from literary works, was carefully designed to avoid misrepresenting characters or their actions.  All source materials are publicly available.  Furthermore, the evaluation framework focuses on assessing LLMs' simulation capabilities rather than generating real-world actions.  We acknowledge the potential for bias in the literary sources used and emphasize the importance of responsible use of this benchmark and its findings. 
The datasets used in our experiment are publicly released and labeled through interaction with humans in English. In this process, user privacy is protected, and no personal information is contained in the dataset. The scientific artifacts that we used are available for research with permissive licenses. And the use of these artifacts in this paper is consistent with their intended use. Therefore, we believe that our research work meets the ethics of ACL.

\section*{Acknowledgements}
This paper is supported by NSFC project 62476009.

\bibliography{custom}

\appendix

\section{Appendix}
\label{sec:app}

\subsection{Data Analysis}
\label{app:data_analysis}
\subsubsection{Data Statistics}

\begin{figure}[h]
    \centering
    \includegraphics[width=1\linewidth]{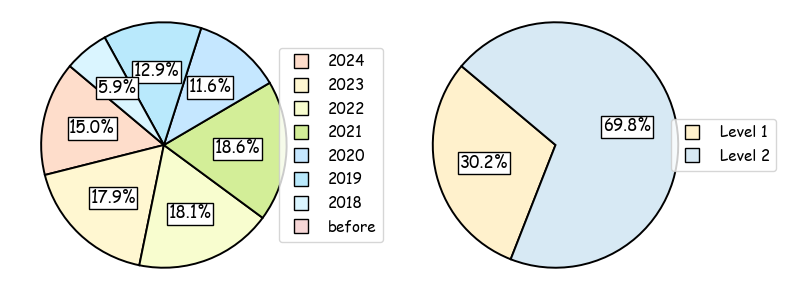}
    \caption{The left figure illustrates the publication year distribution of literary works used as raw material from \datasetname. The right figure shows the proportion of Level 1 and Level 2 behaviors within the overall dataset.}
    \label{fig:bing}
\end{figure}

\begin{table}[h]
\centering
\resizebox{0.5\textwidth}{!}{%
\begin{tabular}{lc}
\toprule
Statistic & Value \\
\midrule
Language & English \\
Source & Novels, Biographies \\
Total Behaviors & 15846 \\
Total Persons & 1001 \\
Max Behavior Nodes & 20 \\
Min Behavior Nodes & 10 \\
Average Behavior Nodes & 15-16 \\
Average Behavior Token Length & 18.48 \\
Style Categories & 25 \\
Genre Categories & 65 \\
Setting Categories & 536 \\
Theme Categories & 115 \\
\bottomrule
\end{tabular}%
}
\caption{Descriptive Statistics of the Behavior Dataset}
\label{tab:behavior_stats} 
\end{table}

\begin{table}[h]
\centering
\small
\resizebox{0.5\textwidth}{!}{%
\begin{tabular}{lc}
\toprule
Statistic & Value \\
\midrule
Style Categories & 25 \\
Genre Categories & 65 \\
Setting Categories & 536 \\
Theme Categories & 115 \\
Lexile Level Categories & 112 \\
Recommended Reading Age Categories & 31 \\
Max Rating Number & 3,380,109 \\
Min Rating Number & 36 \\
Max Ratings Score & 4.7 \\
Min Ratings Score & 3.3 \\
\bottomrule
\end{tabular}%
}
\caption{Book Diversity Statistics}
\label{tab:book_diversity} 
\end{table}

We collect book-related metadata from SuperSummary, including various categorical tags such as Style, Genre, Setting, Theme, Lexile Level, and Recommended Reading Age from SuperSummary. 
Additionally, we record the number of ratings to the rating scores from Goodreads. 
As shown in Table \ref{tab:book_diversity}, the dataset demonstrates a diverse and comprehensive representation of books.

For specific tags, the most prevalent Styles include \textit{Emotional, Mysterious, and Contemplative}. 
Common Genres featured in the dataset are \textit{Historical Fiction, Modern Classic Fiction, Romance, and Mystery \& Crime Fiction}.
Frequently occurring Settings include \textit{Contemporary, 2010s, and United States}. 
The dataset also highlights recurring Themes such as \textit{Family, Love, Friendship, and Race}. 
The most frequently occurring levels include 830L, 780L, HL690L, and 740L.
Additionally, the Recommended Reading Age ranges from \textit{7 to 18+}.
Together, these attributes demonstrate the richness and diversity of the dataset, ensuring a broad representation of literary elements across various themes, genres, reading levels, and target audiences.

Considering data sources beyond English-language western literature is crucial for mitigating bias. 
While our benchmark and the literature used for data construction are in English, our data sources extend beyond English-language western literature. 
They also encompass English translations from various other languages such as Journey to the West. 
We have compiled statistics on the authors' nationalities, revealing contributions from 54 countries, including Canada, USA, UK, New Zealand, India, Japan, Australia, Nigeria, Northern Ireland, France, Ireland, Mexico, Malaysia, Iran, Ethiopia, Zimbabwe, Chile, Argentina, Palestine, Scotland, Ghana, Tanzania, Hungary, Turkey, Spain, Sierra Leone, Indonesia, Pakistan, Afghanistan, Egypt, Jamaica, Italy, China, Lebanon, Vietnam, Thailand, Sri Lanka, Netherlands, Philippines, South Sudan, Yemen, Cuba, North Korea, Switzerland, South Africa, Germany, Democratic Republic of the Congo, Mauritania, North Korea, Liberia, Iceland, Polish British dual nationality, Austria, and Russia, thus ensuring a degree of cultural diversity.

\subsubsection{Impact on Real-World Digital Twin Applications}
Initially, our research focuses on evaluating the general capability of LLMs as human digital twins, a foundational step crucial for applications in specialized fields~\cite{park2023generativeagentsinteractivesimulacra}. Once LLMs can effectively simulate the continuous behaviors of diverse individuals, they will possess broad capabilities to replicate specific individuals across various domains, including healthcare, personalized services, education, and psychological counseling.
Moreover, we extracted persona data from biographies and novels, encompassing a broad range of professional profiles (including doctors, service industry professionals, and other diverse occupations). Biographies provide us with real-world behavioral samples of authentic individuals across various fields. While novels incorporate elements of imagination, they also reflect the authors' realistic observations and experiential insights into these professions~\cite{mckillop1958rise,posen2018doctor}. Thus, this comprehensive dataset will serve as a robust research foundation and provide crucial data support for future real-world digital twin applications.

\subsubsection{Benchmark Features}

(1) Grounding data in real-world source (e.g., literature, expert analysis), (2) integrating rich persona information, including detailed profiles, histories, and coherent context-behavior chains, (3) emphasizing behavior continuity, highlighting the importance of behavior chains over isolated behaviors in independent scenarios—an aspect that is critical for digital twin applications, and (4) employing distinct task designs, including multi-choice for straightforward and accurate quantification and generation for granting the model greater autonomy.


\subsection{Annotation Details}
\label{app:annotation}
\subsubsection{Behavior Check Details}

Low-quality behaviors primarily exhibit naivety and similarity. To ensure the high quality of the extracted behaviors, we employed a multi-round filtering process combining LLMs and human input.

\paragraph{Iteration 1} Real-time filtering during behavior chain extraction. Since each behavior extraction is built upon previous extracted behavior, each new behavior underwent a similarity check against preceding behaviors. If a similar behavior was identified, re-extraction was performed.

\paragraph{Iteration 2} LLM-based naivety (meaningless) filtering. Following the collection of each persona's behavior chain, we implemented an LLM-based naivety check (e.g., daily behaviors that neither reflect the protagonist's personality/emotion/motivation nor are relevant to the context), Each chain was evaluated three times, and nodes consistently flagged as meaningless were filtered, resulting in a 9.84\% behavior reduction.

\paragraph{Iteration 3} Human annotation for behavior node validity. We recruited annotators to screen the behavior node according to a guideline, marking all invalid behavior nodes. This process resulted in a 5.54\% reduction. Annotations were performed in pairs, yielding an average Kappa coefficient of 0.8551. Only behavior nodes agreed upon by both annotators were retained for dataset quality. We authors then conducted a final, comprehensive review of all behavior nodes to guarantee dataset quality.

\paragraph{Iteration 4} Final authors-led comprehensive behavior node screening. We authors conducted a final comprehensive screening of the behavior nodes, ensuring that all behavior nodes were valid, we filtered out less than 1\% of the behavior nodes.


\subsubsection{Guidelines}
\label{app:guildline}
\begin{tcolorbox}[title = {Behaviors Meaningful Check Guidelines }, breakable]

Behaviors that generally need to be removed typically have the following characteristics:\\
\\
\textbf{Overly Common and Mundane Behaviors:} Many behaviors are ordinary actions that humans might perform in any situation (e.g. looking, listening, walking, and saying simple information).\\
\textbf{Simple Interpersonal Behaviors:} Some behaviors are merely simple interactions between characters (e.g. handing something over or answering a question, which themselves lack significant meaning).\\
\textbf{Narrative "Breathing" Behaviors:} Certain behaviors (e.g. Violet announcing her arrival or Beth asking for a beer) might be for narrative flow or to make the scene feel more realistic, but they are not central to the story's driving force.\\
\textbf{Everyday Chores and Simple Reactions:} Many behaviors are very ordinary daily actions, (e.g. turning to see a friend, wagging a finger, interjecting in a conversation, opening one's mouth to respond, announcing arrival, asking a question, calling out a name, looking at someone, listening to an explanation, sitting down, looking at something, listening to a narration, taking something, nodding, putting something down, and looking at someone.) These behaviors themselves lack particular importance.\\
\textbf{Passive Reactions Lacking Emotional/Motivational Drive:} Behaviors are merely instinctive responses to the environment and do not reflect character traits or goals (e.g. frowning due to noise, shrugging shoulders due to cold).\\
\textbf{Technical/Procedural Behaviors:} Simply describing operational processes or physical movement without metaphor or plot advancement (e.g. turning on a machine, tying shoelaces, turning a page of a book).\\
\textbf{Repetitive Behaviors:} Behaviors that repeat the previous or next behavior.\\
\textbf{Homogeneous Group Behaviors:} Individual behaviors in group scenes tend to be the same, without using differentiated behaviors to shape characters (e.g. everyone at a dinner party raising their glasses and smiling, with no one acting differently).\\
\textbf{Transitional Behaviors for Time/Space Jumps:} Simply marking the passage of time or a change in location without carrying narrative function (e.g. He walked through the hallway).\\
\textbf{Culturally/Contextually Default Behaviors:} Routine behaviors in specific cultures or settings that do not need emphasis (e.g. cutting steak in a Western restaurant or passing documents in a meeting room).\\
\textbf{Accumulation of Sensory Description:} Piling up descriptions of the five senses to ``create atmosphere'' without linking them to the characters' psychology (e.g. She smelled jasmine).\\
\textbf{Redundant Dialogue Tags:} Adding meaningless behaviors to avoid repeating the word "said" (e.g. ```I agree,' he scratched his arm'').\\
\textbf{Zero State-Changing Behaviors:} Performing actions that do not change the state of the character or the environment (e.g. ``He wiped his sword,'' and the blade neither became brighter nor revealed a hidden inscription).

\end{tcolorbox}


\subsection{More Results Analysis}
\label{sec:app_result}

\subsubsection{Error Analysis}
we have thoroughly discussed the typical failure modes of models from a macro perspective in Section 4.3, (Analysis). For example, we observed: 1) deficiencies in simulating non-critical behaviors, 2) the transient gains of In-Context Learning and the chained decay of LLMs' simulation capability (meaning that as the behavior chain lengthens, the model's simulation becomes increasingly inaccuracy), and 3) a tendency for models to favor objectively optimal behaviors over those that best fit the character(particularly pronounced in characters with negative personality traits).
Additionally, we provide the following typical error patterns: 1) Omission of history (Knowledge Deficiency, Social Interaction Failure), 2) Out of context (Event Misinterpretation, Temporal Inconsistency, Emotional Distortion), 3) Out of persona profile (Character Misinterpretation, Motivational Misunderstanding).

\subsubsection{GPT-4o as evaluator}
Traditional metrics like ROUGE, BLEU, and sentence similarity, which typically compare generated text against the ground truth, are not well-suited for our behavior generation setting. 
As we aim to provide the target model with greater generative freedom and effectively address the probable one-to-many problem—where, given a context/persona/history of behavior, multiple behaviors may be reasonable. 
Therefore, we adopted GPT-4o as our evaluator because it allows us to evaluate the generated behaviors based on their alignment with the given context and persona, rather than strict adherence to ground-truth.
As for the validity of using LLMs as judges, it is an increasingly accepted and prevalent paradigm. Many works related to role-play~\cite{tu2024charactereval,wang2024incharacter}, simulation~\cite{yuan2024evaluating,wang2024characterbox}, and other generative tasks that require evaluating the quality of model responses employ LLMs~\cite{li2024leveraginglargelanguagemodels}.

we conducted a small-scale validation of the consistency between human and model judgment. We manually inspected 100 behavior nodes judged as proper (1) and 100 behavior nodes judged as improper (0). 
The human consistency with the model's judgment was 81\% for proper behavior nodes and 87\% for improper behavior nodes.


\subsection{Implementation Details}

\subsubsection{Dataset Construction Details}
\label{sec:app_con}
We employed Claude-3-5-sonnet-20240620 for behavior chain extraction and context refinement. The ChatGPT-4o-latest model was utilized to generate distractor items for each behavior. For other generation tasks with lower computational requirements, we selected GPT-3.5 as the execution model.

\subsubsection{Model Testing Details}
\label{sec:app_exp}
\paragraph{Models for Evaluation} All evaluated models were post-interaction versions. The history provided to the LLM consisted of summaries from all preceding chapters used for behavior chain extraction.


\paragraph{Role-Playing LLMs} We found that existing Role-Playing LLMs~\cite{tu2023characterchat,zhou2023characterglm,li2023chatharuhi,yu2024neeko,wang2024rolellm,shao2023characterllm} primarily focus on the simulation of character dialogue, rather than the simulation of broader character behaviors. 
This is one of the main motivations behind the development of the BehaviorChain dataset: to transcend the scope of dialogue, address the research gaps and deficiencies in continuous human behavior simulation.
This difference in research focus leads to the fact that dialogue-based Role-Playing LLMs are not suitable for evaluation using our dataset.

\paragraph{Data Contamination} To address concern that during the pre-training process, LLMs may have already encountered all the data that can be collected from the real world. Therefore, during our dataset construction and experiments, we implemented multiple measures to minimize the potential impact of data contamination on our evaluation. This includes 1) rewriting all data originating from the web and 2) replacing all character name and location entities. Through this meticulous data processing, even if the model has encountered similar plots during pre-training, it is difficult for it to directly apply memorized text. We also include a 3) hard set that contains data published after 2024, which the models are highly unlikely to have been trained on. This fundamentally eliminates the possibility of the model encountering this data during pre-training. For users concerned about data leakage, they can refer to the model's performance on the hard set to assess its ability to simulate continuous behaviors. 4) Additionally, the limited performance of models on our dataset indicates that even if there is a data contamination issue in our task, it is very limited.

\paragraph{Other details} For conciseness, all analyses in Section \ref{sec:analysis} are based on the multiple-choice task framework. 
The single experimental data presented in \ref{sec:analysis} exclusively comprises outputs from the Llama-3.1-70B-Instract model as the target model.
For inputs that exceed the model's context length, we truncated the history from the oldest to the newest.

\subsubsection{Distractors generation details}
Writing and validating distractors is a complex task. We have invested great effort in an iterative trial-and-error process for generating distractors. The preparatory work we conducted prior to large-scale generation has greatly contributed to ensuring their validity.
1) Prompt design: we conducted a continuous trial-and-error process of prompt design with human involvement. The process involved iterative testing as follows:

START: A prompt was first tested on a single behavior node → Distractors of the behavior → Human evaluation

→ Passed human evaluation → Consistently tested the prompt throughout the entire behavior chain → Multiple distractors of each behavior in the chain → Human evaluation

→ All distractors passed human evaluation → consistently tested the prompton more behavior chains (using 5 samples from classic personas as test cases) → Multiple distractors of each behavior in each chain → Human evaluation 

→ All distractors passed human evaluation → Final prompt

(If any human evaluation is not passed during the above stages, we will return to the prompt design phase. )

We finalized the potential prompt after the process described above (in which we also found that using character personality as an anchor is a highly effective strategy for generating distractors). We then scaled up by applying this prompt to generate distractors for nearly 400 behavior nodes across 25 classic personas. All distractors were manually reviewed by the authors. Excluding ground-truth cases with inherent issues (e.g., meaningless behaviors), the generated distractors were consistently distinct from the ground-truth behaviors and reflected different personality traits. These results demonstrate that our prompt strategy effectively produces valid distractors.


\subsection{Prompt}
\label{sec:app_prompt}

\begin{tcolorbox}[title = {First Behavior Generation Prompt}, breakable]

You are an expert in Narrative Analysis and Character Behavior Extraction.\\
Please extract the MOST KEY behavior of \{character\} FROM <Paragraphs>.\\
The behavior should have a significant impact on the development of the storyline, reflect character characteristics or emotions.\\

Ensure that the KEY behavior is an objective statement, clearly stated without any vague expressions.\\
DO NOT add subjective interpretations or inferences about the character's behaviors. ONLY describe the KEY behavior itself. DO NOT mention the result in the KEY behavior.\\
Use your own words instead of quoting the original text.\\
DO NOT repeat or imitate <Previous Key Behavior>.\\

The KEY behavior should have a significant impact on the development of the storyline, the characterization of the characters, and the expression of the theme.\\
Ensure that the key behavior is an objective statement, clearly stated without any vague expressions.\\
DO NOT add subjective interpretations or inferences about the character's behaviors. Only describe the behavior itself.\\

The format of your response should be: \{"key behavior": ""\}.\\

<Paragraphs BEGIN>\\
\{parts[0]\}\\
<Paragraphs END>\\

If the behavior of \{character\} cannot be extracted, output "" ONLY.

\end{tcolorbox}


\begin{tcolorbox}[title = {Next Behavior Generation Prompt}, breakable]
You are an expert in Narrative Analysis and Character Behavior Extract.\\
Below, I will provide you with <Previous Paragraphs>, <Previous Key Behavior> extracted from <Previous Paragraphs> and <Current Paragraphs>. \\

Please summary the scene change and plot development detailly and naturally after the <Previous Key Behavior> according to the <Previous Paragraphs> and <Current Paragraphs> I give you. The summary should start with "\{examples[-1]['key behavior']\} After that, ".\\

After the summary, you should extract the MOST KEY behavior of \{character\} FROM <Current Paragraphs>, describe in more than 10 words.\\
The behavior should be a non-meaningless behavior taken spontaneously by \{character\}.\\
The behavior should have a significant impact on the development of the storyline or reflect character characteristics or emotions.\\
Ensure that the behavior is objective statements and state the behavior clearly and do not use any vague expressions. \\
DO NOT add subjective interpretations and inference about the character's behaviors. ONLY describe the behavior itself. \\
DO NOT mention the result in the behavior.\\
Ensure use your own words instead of quoting the original text. \\
DO NOT repeat or imitate <Previous Key Behavior>. \\

Please provide a REVISED summary of the scene change and plot development that occurred before the behavior you extracted from <Current Paragraphs>, making sure not to reveal any information about the behavior. Delete the behavior and subsequent plots, and keep only the plots before the behavior.\\
The REVISED summary should end with "After this or in response to this, what behavior did \{character\} take?"\\

<Previous Paragraphs BEGIN>\\
\{marge part[-1]\}\\
<Previous Paragraphs END>\\

<Previous Key Behavior BEGIN>\\
\{examples[-1]['key behavior']\}\\
<Previous Key Behavior END>\\

<Current Paragraphs BEGIN>\\
\{parts[i]\}\\
<Current Paragraphs END>\\

The format of your response should be \{"summary": "","key behavior":"","new summary":""\}.\\
Ensure that the behavior you extract is taken by \{character\} at this moment, rather than behavior of others or past behavior (in <Paragraphs>, \{character\}probable in the first person).\\
If the behavior of \{character\} cannot be extracted, output "None" ONLY.

\end{tcolorbox}

\begin{tcolorbox}[title = {Similirity Check}, breakable]

Please determine whether the following two behaviors refer to the same behavior:\\
Behavior 1: \{sentence1\}\\
Behavior 2: \{sentence2\}\\
If there is a strong possibility that the two behaviors refer to the same behavior, please output 1; otherwise, output 0, Make sure you give me 0/1.
\end{tcolorbox}

\begin{tcolorbox}[title = {Context Refine}, breakable]
<Context Begin>\\
\{relevant part\}\\
<Context End>\\

<Behavior Begin>\\
\{key behavior\}\\
<Behavior End>\\

Your task is to refine the <Context> according to the following requirements:\\

1. If the <Context> explicitly or implicitly suggests the active behaviors of \{character\} in the <Behavior> or discloses the result/reactions of others caused by <Behavior>, delete these from <Context>. Any references to \{character\}' emotions, feelings, psychological states, or internal conflicts should be eliminated from <Context>.\\

2. If < Behavior> include \{character\}'s reaction/response to event/situation/others behavior, then that event/situation/others behavior should to be described intactly and directly in the end of <Context>.\\

3. If the <Behavior> includes any elements such as contextual conditions or encounters rather than purely active behavior of \{character\}, integrate these elements into the <Context>. Pay attention to the clauses in <Behavior> as it often contain contextual information, but do not include it in <Context> if it happens after \{character\}'s behavior.\\

4. Output the refined <Context> directly without other note.

\end{tcolorbox}

\begin{tcolorbox}[title = {Distractor Generation}, breakable]
<Context Start>\\
\{examples[i]['summary refined']\}\\
<Context End>\\

The original subsequent behavior was "\{behavior\}".\\
Estimate what personality trait does this reflect in brief words, and generate 3 behaviors that different personality traits would exhibit, answered in JSON format.\\
\{\\
"original behavior traits":"",\\
"difference": [\\
\{\\
 "trait": "",\\
 "behavior": ""\\
 \},\\
\{\\
 "trait": "",\\
"behavior": ""\\
 \},\\
\{\\
 "trait": "",\\
 "behavior": ""\\
 \}\\
]\\
\}

\end{tcolorbox}

\begin{tcolorbox}[title = {Level Define}, breakable]
<Summary BEGIN>\\
\{summary l\}\\
<Summary END>\\
Please check whether The following behaviors are implied in the summary.\\
In your reply, retain the sequence numbers. The number of key behaviors must be considerably fewer than \{int(n/2)\}.\\

<behaviors BEGIN>\\
\{sentences\}\\
<behaviors END>

\end{tcolorbox}

\subsection{Data instance}
\label{sec:app_instance}
\begin{tcolorbox}[title = {Profile}, breakable]
\{
  "Name": "Jay Gatsby",\\
  "Personality Traits": [
    "Charismatic",
    "Mysterious",
    "Obsessive",
    "Romantic",
    "Wealthy",
    "Idealistic"
  ],\\
  "Motivations and Goals": [\\
    "To reunite with his former lover, Daisy Buchanan",\\
    "To achieve a high social status and wealth",\\
    "To recapture the past and fulfill his ideal vision of life with Daisy"\\
  ],\\
  "Significant Background Events": [\\
    "Born James Gatz to a poor farming family in North Dakota",\\
    "Changed his name to Jay Gatsby and reinvented himself as a wealthy socialite",\\
    "Amassed his fortune through questionable means",\\
    "Became known for his lavish parties at his mansion in West Egg"\\
  ],\\
  "Relationships": \{\\
    "Daisy Buchanan": "Former lover, whom Gatsby is still deeply in love with",\\
    "Tom Buchanan": "Daisy's husband and Gatsby's rival",\\
    "Nick Carraway": "Narrator of the story and Gatsby's neighbor and friend",\\
    "Jordan Baker": "A professional golfer and friend of Daisy, whom Gatsby has a brief romantic interest in",\\
    "George Wilson": "A mechanic and owner of a garage, indirectly involved in Gatsby's downfall"\\
  \},\\
  "Additional Details": \{\\
    "Occupation": "Businessman with mysterious sources of wealth",\\
    "Social Status": "Wealthy and influential, but not born into old money",\\
    "Hobbies": "Throwing extravagant parties, collecting expensive art and cars",\\
    "Residence": "A grand mansion in West Egg, New York"\\
  \}\\
\}\\
\end{tcolorbox}

\begin{tcolorbox}[title = {History}, breakable]
Summary of Chapter 1 \\
Nick, a Yale graduate and World War I veteran, moves to West Egg, Long Island, to work as a bond salesman.  He rents a small house next to the opulent mansion of his mysterious neighbor, Jay Gatsby.

West Egg is separated from the more fashionable East Egg by the Long Island Sound.  Across the water, Nick's cousin Daisy lives with her wealthy and imposing husband, Tom Buchanan, whom Nick knew at Yale.  Tom's wealth and social standing are evident in their "Georgian Colonial mansion."  Also present at their home during Nick's visit is Jordan Baker, a friend of Daisy's and a well-known golfer.

The chapter highlights the Buchanans' superficial and privileged lifestyle.  During dinner, Tom reveals his racist views by referencing a book espousing white supremacist theories, claiming the Nordic race is responsible for civilization.  This disturbs Nick, but no one challenges Tom's comments.

The dinner is interrupted by a phone call, hinting at Tom's infidelity.  Jordan informs Nick that Tom is having an affair with a woman in New York City.  The tension is palpable, but Daisy attempts to deflect attention.  The awkwardness of the situation is further amplified by another phone call.

As the evening progresses, Nick learns that Jordan Baker is a famous golfer with a somewhat scandalous reputation.  He also discovers that Daisy and he, despite being related, are not close.  The Buchanans tease Nick about a rumor of a broken engagement, which he denies.

Returning home, Nick sees Gatsby standing outside his mansion.  Nick considers inviting him over but hesitates, sensing something enigmatic about Gatsby's presence.
\\Summary of Chapter 2 \\
Chapter 2 delves into the "valley of ashes," a desolate area between West Egg and New York City, where industrial ashes are dumped.  Presiding over this wasteland is the faded billboard of Dr. T.J. Eckleburg, featuring giant, spectacled eyes.

This grim setting is significant because it's where George Wilson's struggling auto shop and his wife Myrtle, Tom's mistress, are located.  Tom, on his way to New York City with Nick, stops at Wilson's garage, ostensibly to discuss a car deal.  The real purpose, however, is to rendezvous with Myrtle. Nick finds Myrtle physically unattractive, despite her apparent allure for Tom.

The group, leaving George behind, proceeds to New York City, impulsively buying a puppy along the way.  They gather at a small apartment Tom keeps for his affair.  The other guests include Myrtle's sister, Catherine, and the McKees, neighbors from the building. Mr. McKee is a mediocre photographer.

The atmosphere becomes increasingly tawdry and tense.  A critical moment occurs when Myrtle mentions Daisy's name, provoking Tom to violently strike her, breaking her nose.

Nick, claiming to be drunk (only the second time in his life), leaves the apartment with Mr. McKee amidst the chaos of tending to Myrtle's injury.  He experiences a blackout and then finds himself at Mr. McKee's bedside, observing his amateurish photographs.  Nick manages to escape the apartment and waits at the train station for the 4:00 a.m. train home.
\\Summary of Chapter 3 \\
Chapter 3 depicts one of Gatsby's extravagant parties through Nick's eyes.  Nick receives a formal invitation, a stark contrast to the casual attendance of most guests, who often don't even know Gatsby.  At the party, Nick reconnects with Jordan Baker, who explains she enjoys large gatherings for the anonymity they offer.  A humorous anecdote highlights Gatsby's generosity: he replaced a dress torn at a previous party with an expensive new one.

Rumors about Gatsby's background and wealth circulate among the guests, ranging from accusations of being a German spy to claims of him being a war hero or even a murderer.  While searching for Gatsby, Nick encounters a drunken "owl-eyed" man in Gatsby's impressive library, who is astonished to discover the books are real.  Nick eventually meets Gatsby, but their conversation is cut short by a phone call.  Observing Gatsby later, Nick is struck by his charismatic smile and the contrast between his apparent sobriety and the revelry of his guests.  Gatsby then takes Jordan aside for a private conversation, leaving her visibly shaken.  As the party ends, a drunken car crash involving the owl-eyed man underscores the reckless atmosphere.

Nick then reflects on his experiences, noting that his time in West Egg isn't solely filled with social events but also with work.  He expresses his growing affection for New York City, despite feelings of loneliness, and his burgeoning relationship with Jordan.  He recounts a story about Jordan cheating in a golf tournament, a scandal that, though dismissed, hints at her dishonesty.  Nick rationalizes her behavior as a consequence of her desire for independence clashing with her need for protection.

The chapter concludes with Nick and Jordan becoming romantically involved.  Nick reveals he ended a relationship in the Midwest to pursue Jordan and emphasizes his own sense of honesty.
\\Summary of Chapter 4 \\
Chapter 4 of *The Great Gatsby* opens with a Sunday morning at Gatsby's mansion, where the usual rumors about his past continue to circulate. Nick briefly catalogs the diverse and affluent guests who frequent Gatsby's parties.

One day in late July, Gatsby takes Nick to lunch in his luxurious car.  During the drive, Gatsby seems overly concerned with Nick's opinion of him and his vehicle. He then directly addresses the rumors about his background, claiming to be the son of wealthy, deceased Midwestern parents.  However, he gives the contradictory answer of San Francisco when asked where in the Midwest.  He further elaborates, stating that he inherited his parents' fortune, lived lavishly in Europe, and then became a war hero in World War I, receiving numerous medals.

To support his claims, Gatsby shows Nick a photo of himself at Oxford and a medal from Montenegro.  These items, along with Gatsby's ability to evade a speeding ticket by showing a Christmas card from the police commissioner, convince Nick of the truth of Gatsby's stories.

They have lunch with Meyer Wolfsheim, a shady character who, according to Gatsby, fixed the World Series. Wolfsheim's bizarre cufflinks, made of human molars, further emphasize his connection to the criminal underworld.  Tom Buchanan briefly joins them, but Gatsby disappears abruptly.

Later, Jordan reveals to Nick that she met Gatsby in Louisville in 1917, when Daisy and Gatsby were together.  Jordan recounts how Daisy almost ran off to New York with Gatsby before he went to war, and how she was later deeply unhappy before her wedding to Tom.  Jordan then reveals the reason for her private conversation with Gatsby at the party: Gatsby wants to reunite with Daisy, and he needs Nick's help to arrange a meeting.
\\Summary of Chapter 5 \\
Chapter 5 details the long-awaited reunion between Gatsby and Daisy at Nick's cottage.  Before Daisy arrives, Gatsby's feigned disinterest crumbles as he obsessively prepares, even having Nick's lawn manicured.  Daisy's initial question about the purpose of the invitation and whether Nick harbors romantic feelings for her is met with a reference to *Castle Rackrent*.

The meeting begins awkwardly, and Nick tries to give them space. However, Gatsby, clearly anxious, repeatedly seeks Nick's advice, prompting Nick to reprimand him for his behavior.  Gatsby's reaction suggests he's offended by Nick's criticism.

They then move to Gatsby's mansion. Nick briefly describes the mansion's original owner, a wealthy brewer who, according to rumors, wanted the surrounding area to resemble a medieval village, an idea Nick finds inherently un-American.  The only person present in the vast house is Klipspringer, a freeloading "boarder."  Gatsby proudly displays his extensive and expensive collection of shirts, a display that moves Daisy to tears.  Nick notices a picture of a man on a yacht, whom Gatsby identifies as Dan Cody, his former mentor and close friend.

Gatsby points out the green light on Daisy's dock, visible from his house in clear weather.  After touring the mansion, they again encounter Klipspringer, whom Gatsby compels to play the piano.  Finally, Nick leaves Gatsby and Daisy alone, implying the possibility of their rekindling their past romance.
\\Summary of Chapter 6 \\
Chapter 6 begins with a reporter's visit to Gatsby, seeking a statement regarding vague, unsubstantiated rumors.  This prompts Nick to recount Gatsby's true origins, as told by Gatsby himself.  James Gatz, born to an unremarkable family in South Dakota, reinvented himself as Jay Gatsby after meeting Dan Cody on Lake Superior.  Feeling out of place in his hometown, Gatz left to pursue a grander future.  He encountered Cody's yacht and, through his resourcefulness, became Cody's personal assistant and was even named a potential heir. However, after Cody's death, his mistress, Ella Kaye, intervened, preventing Gatsby from inheriting anything.\\
Weeks after Gatsby and Daisy's reunion, Nick unexpectedly encounters Tom Buchanan at Gatsby's mansion. Tom and two acquaintances have stopped by during a horseback riding trip. Gatsby's overly eager hospitality towards Tom is noticeable. When Tom and his group leave for another gathering, Gatsby almost insists on joining them, despite their clear disinterest.

Later, Tom and Daisy attend one of Gatsby's parties. Nick observes a palpable tension, viewing the West Egg revelry through Daisy's East Egg perspective.  Daisy is clearly repulsed by the spectacle.  During the party, Tom fuels the rumors of Gatsby's bootlegging activities, disparaging him as "newly rich."  Daisy half-heartedly defends Gatsby, claiming the guests are uninvited and that Gatsby is too gracious to turn them away.  Tom openly flirts with other women, while Daisy, in a strange moment, invites Nick to kiss her, an invitation he declines.

After Tom and Daisy depart, Nick stays late. Gatsby confides in Nick about Daisy, revealing his desire for her to leave Tom and be with him. Gatsby then wistfully recalls the pivotal moment, five years prior, when he first kissed Daisy, a memory he cherishes as a defining moment in his life.

\end{tcolorbox}

\begin{tcolorbox}[title = {Behavior Chain}, breakable]
<context 1>\\
Gatsby dismissed every servant in his house and replaced them with new ones who did not interact with the local community. After that, rumors spread in the village about the new servants, and Gatsby explained to Nick that he hired people recommended by Wolfsheim who wouldn't gossip. The next day was extremely hot, and Nick traveled to the Buchanans' house by train. After this or in response to this, what behavior did Jay Gatsby take?
\\<key\_behavior 1>\\
Gatsby called Nick on the phone to invite him to lunch at Daisy's house the next day.
\\<distractors 1>\\
Gatsby shows up unannounced at the Buchanans' house, insisting on seeing Daisy immediately despite the potential awkwardness.\\
Gatsby avoids contact with Nick or Daisy altogether, choosing instead to observe the situation from a distance without direct involvement.\\
Gatsby openly declares his feelings for Daisy in public, ignoring the potential consequences and drawing attention to himself.\\
\\<context 2>\\
Gatsby called Nick on the phone to invite him to lunch at Daisy's house the next day. After that, Nick arrived at the Buchanans' house on a sweltering hot day. The butler answered the phone, and Nick and Gatsby were directed to the salon where Daisy and Jordan were resting on a couch. Tom's voice could be heard on the telephone in the hall, apparently having a heated conversation about selling a car. Tom then entered the room and greeted Gatsby with concealed dislike. The room had a crimson carpet. After this or in response to this, what behavior did Jay Gatsby take?
\\<key\_behavior 2>\\
Gatsby stood in the center of the crimson carpet and gazed around with fascinated eyes.
\\<distractors 2>\\
Gatsby strode toward Tom with a firm handshake and a polite but assertive comment about the weather.\\
Gatsby fidgeted with the edge of his jacket while avoiding making direct eye contact with anyone in the room.\\
Gatsby crossed his arms, gave Tom a steely glare, and responded curtly to any pleasantries.\\
\\<context 3>\\
Gatsby stood in the center of the crimson carpet and gazed around with fascinated eyes. After that, the group moved to the veranda upon Tom's suggestion. They observed the green Sound where a small sail was moving slowly in the heat. After this or in response to this, what behavior did Jay Gatsby take?
\\<key\_behavior 3>\\
Gatsby raised his hand and pointed across the bay to show the location of his house
\\<distractors 3>\\
Gatsby remained silent, simply nodding in acknowledgment of the view without drawing attention to his house.\\
Gatsby spoke poetically about the beauty of the green Sound and the sailboat, avoiding mention of his house entirely.\\
Gatsby excused himself quietly to stand apart, gazing at the water in introspection.\\
\\<context 4>\\
Gatsby raised his hand and pointed across the bay to show the location of his house. After that, the group moved to have lunch in the darkened dining room where they drank cold ale. The atmosphere became tense as Daisy expressed her distress about the heat and confusion, suggesting they all go to town. Tom and Gatsby engaged in small talk about garages and stables, while Daisy persistently pushed for the idea of going to town. When Daisy asked again who wanted to go to town After this or in response to this, what behavior did Jay Gatsby take?
\\<key\_behavior 4>\\
Gatsby's eyes floated toward Daisy when she asked about going to town, and he stared at her, maintaining eye contact
\\<distractors 4>\\
Gatsby casually shrugged and glanced out the window, showing disinterest in Daisy's suggestion.\\
Gatsby immediately stood up and said, 'Let’s go right now!' without waiting for the group’s consensus.\\
Gatsby hesitated, thoughtfully considering whether going to town was practical given the heat and tension in the room.\\
\\<context 5>\\
Gatsby's eyes floated toward Daisy when she asked about going to town, and he stared at her, maintaining eye contact. After that, no one immediately moved to leave. Tom became increasingly agitated at the group's hesitation, and his hand trembled as he finished his drink. Daisy tried to delay their departure by suggesting they smoke first, but Tom dismissed her suggestion. The women went upstairs to prepare while the three men waited outside in the heat. A tense atmosphere developed as they stood shuffling their feet on the hot gravel, with the silver moon already visible in the western sky. Tom stood there with them. After this or in response to this, what behavior did Jay Gatsby take?
\\<key\_behavior 5>\\
Gatsby attempted to speak to Tom but stopped himself, then asked about Tom's stables with visible effort
\\<distractors 5>\\
Gatsby confronted Tom directly, making a pointed comment about Daisy's choices or feelings, his voice unwavering.\\
Gatsby abruptly suggested they leave for town without consulting anyone, his restlessness showing in his tone and actions.\\
Gatsby smirked and made a sly, cutting remark about Tom's stables, using it as a veiled critique of Tom's character.\\
\\<context 6>\\
Gatsby attempted to speak to Tom but stopped himself, then asked about Tom's stables with visible effort. After that, the group prepared to leave for town. Daisy and Jordan went upstairs to get ready while the men waited outside. Tom expressed his displeasure about going to town, and went inside to get whiskey. The narrator made a comment about Daisy's voice, and Gatsby described it as being 'full of money.' Tom returned with a bottle of whiskey, followed by Daisy and Jordan. Tom suggested they take his car. After this or in response to this, what behavior did Jay Gatsby take?
\\<key\_behavior 6>\\
Gatsby objected to Tom's suggestion of driving his car, claiming there wasn't much gas.
\\<distractors 6>\\
Gatsby nodded and agreed that Tom's car would be a good choice, showing a willingness to go along with the plan without objection.\\
Gatsby hesitated, closely examining Tom's car, and suggested they double-check everything to ensure the ride would be safe.\\
Gatsby grinned and suggested they all race both cars to town, adding some excitement to the mundane trip.\\
\\<context 7>\\
Gatsby objected to Tom's suggestion of driving his car, claiming there wasn't much gas. After that, Tom insisted there was plenty of gas and suggested stopping at a drugstore if needed. The group then split up, with Tom driving Gatsby's car with Nick and Jordan, while Gatsby and Daisy followed in Tom's coupé. As they drove, Tom expressed suspicion about Gatsby and mentioned conducting a small investigation into his past. They stopped at Wilson's garage for gas, where Wilson revealed his plans to move west with his wife. The group then continued to the city, eventually ending up in a suite at the Plaza Hotel. The conversation became tense as Tom questioned Gatsby's use of the phrase 'old sport'. In the suite, there was a fallen telephone book with its string parted. After this or in response to this, what behavior did Jay Gatsby take?
\\<key\_behavior 7>\\
Gatsby examined the parted string of the fallen telephone book, muttered 'Hum!' in an interested way, and tossed the book on a chair.
\\<distractors 7>\\
Gatsby ignored the fallen telephone book completely, focusing instead on more immediate concerns in the conversation.\\
Gatsby picked up the fallen telephone book, carefully retied the string, and placed it neatly back where it belonged.\\
Gatsby glanced briefly at the telephone book, shrugged, and dismissed it as unimportant without further reaction.\\
\\<context 8>\\
Gatsby examined the parted string of the fallen telephone book, muttered 'Hum!' in an interested way, and tossed the book on a chair. After that, Tom sharply criticized Gatsby's habit of saying 'old sport,' leading to tension in the room. Daisy attempted to diffuse the situation by requesting ice for mint juleps. The group then heard a wedding march from downstairs, which prompted a discussion about Daisy's wedding and a mysterious guest named Biloxi. As the conversation about Biloxi's conflicting stories continued, Tom suddenly turned to Gatsby with a pointed question about his education. After this or in response to this, what behavior did Jay Gatsby take?
\\<key\_behavior 8>\\
When questioned about his Oxford education, Gatsby first responded evasively by saying 'Not exactly,' then changed his answer to confirm he went there, and finally specified he attended Oxford for five months in 1919.
\\<distractors 8>\\
Gatsby would directly and calmly state he attended Oxford as part of a special program, elaborating on the details without hesitation.\\
Gatsby would respond assertively, challenging Tom by pointing out the validity of his education and questioning why it matters.\\
Gatsby would shrug and dismiss the question, saying something like, 'I don't see why it matters,' and redirecting the conversation elsewhere.\\
\\<context 9>\\
When questioned about his Oxford education, Gatsby first responded evasively by saying 'Not exactly,' then changed his answer to confirm he went there, and finally specified he attended Oxford for five months in 1919. After that, Gatsby explained that his time at Oxford was part of an opportunity given to some officers after the Armistice, allowing them to attend universities in England or France. This explanation seemed to renew the narrator's faith in Gatsby. Meanwhile, Daisy attempted to lighten the mood by suggesting they make mint juleps, but Tom insisted on asking Gatsby one more question, demanding to know what kind of 'row' Gatsby was trying to cause in his house. After this or in response to this, what behavior did Jay Gatsby take?
\\<key\_behavior 9>\\
Gatsby politely invited Tom to continue with his questioning.
\\<distractors 9>\\
Gatsby openly challenges Tom, questioning Tom's motivations and accusing him of hypocrisy in front of everyone.\\
Gatsby avoids eye contact, becomes visibly agitated, and deflects the question by bringing up a completely unrelated topic.\\
Gatsby abruptly raises his voice, passionately declaring his love for Daisy and confronting Tom about his treatment of her.\\
\\<context 10>\\
Gatsby politely invited Tom to continue with his questioning. After that, The tension in the room escalated as Tom accused Gatsby of causing trouble and challenged the modern attitudes towards family life. Daisy attempted to diffuse the situation, but Tom's anger and prejudice became more apparent. Jordan and Nick reacted with discomfort to Tom's outburst. As the argument intensified, Daisy interrupted, pleading for everyone to leave. Nick agreed and suggested they all go home, but Tom insisted on hearing what Gatsby had to say. After this or in response to this, what behavior did Jay Gatsby take?
\\<key\_behavior 10>\\
Gatsby directly told Tom that his wife doesn't love him.
\\<distractors 10>\\
Gatsby calmly acknowledged the tension and suggested a private conversation with Tom to resolve the issue.\\
Gatsby remained silent and avoided escalating the argument further, allowing others to take the lead.\\
Gatsby expressed understanding of Tom's concerns and reassured him that he had no intention to disrupt the family.\\
\\<context 11>\\
Gatsby directly told Tom that his wife doesn't love him. After that, Gatsby revealed that he and Daisy had been in love for five years without Tom's knowledge. Tom became furious and denied this claim, insisting that Daisy loved him when they married and still loves him now. Tom admitted his own infidelities but claimed he always came back to Daisy. Daisy expressed her disgust at Tom and confronted him about their departure from Chicago. After this or in response to this, what behavior did Jay Gatsby take?
\\<key\_behavior 11>\\
Gatsby walked over and stood beside Daisy to show his support when she confronted Tom about their departure from Chicago
\\<distractors 11>\\
Gatsby stayed silent and remained seated, avoiding direct involvement or confrontation.\\
Gatsby aggressively confronted Tom, escalating the conflict and directly challenging his assertions about Daisy.\\
Gatsby distanced himself from the interaction entirely, stepping away from both Daisy and Tom as they argued.\\
\\<context 12>\\
Gatsby walked over and stood beside Daisy to show his support when she confronted Tom about their departure from Chicago. After that, Gatsby urged Daisy to tell Tom the truth about never loving him, insisting that it would erase their past. Daisy hesitated and looked to Jordan and the narrator for support before reluctantly stating she never loved Tom. Tom questioned her about specific moments in their past, but Daisy refused to engage further. She then turned to Gatsby, acknowledging his presence but expressing frustration at his expectations. After this or in response to this, what behavior did Jay Gatsby take?
\\<key\_behavior 12>\\
Gatsby's eyes opened and closed in response to Daisy's confession of loving both him and Tom.
\\<distractors 12>\\
Gatsby smiles softly at Daisy, nodding with quiet acceptance, and assures her that her feelings won't change his regard for her.\\
Gatsby abruptly interrupts, dismissing Daisy's confusion and insisting they leave together immediately regardless of what she just said.\\
Gatsby steps back, averting his gaze, and calmly acknowledges her feelings, signaling that he would respect whatever decision she makes.\\
\\<context 13>\\
Gatsby's eyes opened and closed in response to Daisy's confession of loving both him and Tom. After that, Tom harshly claimed that Daisy never truly knew Gatsby was alive and mentioned his intimate history with Daisy. After this or in response to this, what behavior did Jay Gatsby take?
\\<key\_behavior 13>\\
Gatsby insisted on speaking to Daisy alone and declared to Tom that he would no longer take care of Daisy.
\\<distractors 13>\\
Gatsby emotionally withdrew, avoiding confrontation altogether, and let Tom and Daisy decide their fate without intervention.\\
Gatsby accused Tom of manipulating Daisy and escalated the verbal conflict, refusing to back down and openly challenging Tom in front of everyone.\\
Gatsby tried to comfort Daisy and suggested that she take her time to figure out her feelings, prioritizing her emotional well-being over himself or Tom.\\
\\<context 14>\\
Gatsby insisted on speaking to Daisy alone and declared to Tom that he would no longer take care of Daisy. After that, Tom accused Gatsby of being a common swindler and revealed that he had conducted an investigation into Gatsby's affairs. Tom exposed Gatsby's involvement in illegal activities, including running drug stores that sold grain alcohol. He also mentioned Gatsby's association with Meyer Wolfsheim and accused him of being a bootlegger. After this or in response to this, what behavior did Jay Gatsby take?
\\<key\_behavior 14>\\
Gatsby calmly responded to Tom's accusations about his illegal activities and association with Meyer Wolfsheim.
\\<distractors 14>\\
Gatsby angrily shouted at Tom, denying the accusations and accusing Tom of trying to smear his name.\\
Gatsby avoided eye contact and quietly admitted to parts of the accusations, retreating from any further confrontation.\\
Gatsby immediately launched into a detailed explanation and justification of his activities, attempting to discredit Tom's claims.\\
\\<context 15>\\
Gatsby calmly responded to Tom's accusations about his illegal activities and association with Meyer Wolfsheim. After that, Tom continued his accusations by bringing up Walter Chase and the betting laws, mentioning how Wolfsheim had intimidated Walter into silence. After this or in response to this, what behavior did Jay Gatsby take?
\\<key\_behavior 15>\\
Gatsby displayed an unfamiliar yet recognizable expression on his face when confronted about his more serious criminal activities
\\<distractors 15>\\
Gatsby smirked confidently and dismissed the accusations as baseless lies, daring Tom to present any real evidence.\\
Gatsby maintained a serene expression, calmly stating that his private affairs were none of Tom's concern.\\
Gatsby suddenly raised his voice, vehemently denying the accusations and accusing Tom of hypocrisy and worse behavior.\\
\end{tcolorbox}

\end{document}